%% file: revise01nov05.tex
\documentclass[english,11pt]{article}

\usepackage{epsfig}
\usepackage{babel,varioref}
\usepackage{a4}
\usepackage{amsmath, amsthm, amssymb}
\usepackage{amsfonts}
\usepackage{latexsym}
\usepackage{graphicx}
\usepackage{texdraw}
\usepackage{multicol}
\usepackage{multirow}
\usepackage{comment}
\usepackage{setspace} 
\usepackage{color} 

\usepackage{epic}

\input{epsf}

\begin{document}

\begin{center} {\Large Submission to \textit{CONSTRAINTS}}
\end{center}
\vspace{1cm}

\begin{center} {\LARGE {Stochastic Constraint Programming: A Scenario-Based Approach}}
\end{center}

\vspace{0.5cm}

\begin{abstract}
To model combinatorial decision problems involving uncertainty and
probability, we introduce scenario based stochastic constraint
programming. Stochastic constraint programs contain both decision
variables, which we can set, and stochastic variables, which follow
a discrete probability distribution. We provide a semantics for
stochastic constraint programs based on scenario trees.
Using this semantics, we
can compile stochastic constraint programs down into conventional
(non-stochastic) constraint programs. This allows us to exploit the
full power of existing constraint solvers. We have implemented this
framework for decision making under uncertainty in stochastic OPL, a
language which is based on the OPL constraint modelling language
[Hentenryck et al., 1999]. To illustrate the potential of this
framework, we model a wide range of problems in areas as diverse as
portfolio diversification, agricultural planning and
production/inventory management. \newline \newline Keywords:
constraint programming, constraint satisfaction, reasoning under
uncertainty
\end{abstract}

\vspace{1cm}

\noindent ${}$ \hspace{2cm} S. Armagan Tarim (primary contact
author)
\newline ${}$ \hspace{2cm} Cork Constraint Computation Centre, \newline ${}$ \hspace{2cm} Department of
Computer Science, \newline ${}$ \hspace{2cm} University College
Cork, Cork, Ireland \newline ${}$ \hspace{2cm} Tel: +353-21-4255411
\newline ${}$ \hspace{2cm} at@4c.ucc.ie,
http://www.cs.york.ac.uk/$\sim$at/
\newline \newline
${}$ \hspace{2cm} Suresh Manandhar \newline ${}$ \hspace{2cm}
Artificial Intelligence Group, \newline ${}$ \hspace{2cm} Department
of Computer Science,
\newline ${}$ \hspace{2cm} University of York, York, UK
\newline ${}$ \hspace{2cm} Tel: +44-1904-432746 \newline ${}$ \hspace{2cm} suresh@cs.york.ac.uk,
http://www.cs.york.ac.uk/$\sim$suresh/
\newline \newline
${}$ \hspace{2cm} Toby Walsh \newline ${}$ \hspace{2cm} National ICT
Australia and School of CSE
\newline ${}$ \hspace{2cm} University of New South Wales, Sydney, Australia
\newline ${}$ \hspace{2cm} Tel:+61-2-93857343 \newline ${}$ \hspace{2cm} tw@cse.unsw.edu.au, http://4c.ucc.ie/$\sim$tw

\newpage

\setcounter{page}{1}

\title{Stochastic Constraint Programming: A Scenario-Based Approach}
\author{
S. Armagan Tarim, \\
Cork Constraint Computation Centre, University College Cork, Ireland    \\
Suresh Manandhar, \\
Department of Computer Science, University of York, United Kingdom   \\
Toby Walsh, \\
National ICT Australia and School of Computer Science and Engineering, \\
University of New South Wales, Australia  \\
}\date{} \maketitle

\begin{abstract}
To model combinatorial decision problems involving uncertainty and
probability, we introduce scenario based stochastic constraint
programming. Stochastic constraint programs contain both decision
variables, which we can set, and stochastic variables, which follow
a discrete probability distribution. We provide a semantics for
stochastic constraint programs based on scenario trees.
Using this semantics, we
can compile stochastic constraint programs down into conventional
(non-stochastic) constraint programs. This allows us to exploit the
full power of existing constraint solvers. We have implemented this
framework for decision making under uncertainty in stochastic OPL, a
language which is based on the OPL constraint modelling language
[Hentenryck et al., 1999]. To illustrate the potential of this
framework, we model a wide range of problems in areas as diverse as
portfolio diversification, agricultural planning and
production/inventory management. \newline \newline Keywords:
constraint programming, constraint satisfaction, reasoning under
uncertainty
\end{abstract}

\begin{center}
\begin{tabular}{ll}
Content areas: & constraint programming, constraint satisfaction, reasoning under uncertainty\\
\end{tabular}
\end{center}

\section{Introduction}

Many 
decision problems contain uncertainty. Data about events in the past
may not be known exactly due to errors in measuring or difficulties
in sampling, whilst data about events in the future may simply not
be known with certainty. For example, when scheduling power
stations, we need to cope with uncertainty in future energy demands.
As a second example, nurse rostering in an accident and emergency
department requires us to anticipate variability in workload. As a
final example, when constructing a balanced bond portfolio, we must
deal with uncertainty in the future price of bonds. To deal with
such situations, \cite{walsh02} proposed an extension of
constraint programming, called {\em stochastic constraint
programming}, in which we distinguish between
decision variables, which we are free to set,
and stochastic (or observed) variables, which follow some
probability distribution. A semantics for
stochastic constraint programs based on policies was proposed
and backtracking and forward checking algorithms to solve such
stochastic constraint programs were presented.

In this paper, we extend this framework to make
it more useful practically. In particular, we permit multiple chance
constraints and a range of different objectives. As each such
extension requires large changes to the backtracking and forward
checking algorithms, we propose instead a scenario based view of
stochastic constraint programs. One of the major advantages of this
approach is that stochastic constraint programs can then be compiled
down into conventional (non-stochastic) constraint programs. This
compilation allows us to use existing constraint solvers without any
modification, as well as call upon the power of hybrid solvers which
combine constraint solving and integer programming techniques. We
also propose a number of techniques to reduce the number of
scenarios and to generate robust solutions. This framework combines
together some of the best features of traditional constraint
satisfaction, stochastic integer programming \cite{rusz03}, and
stochastic satisfiability \cite{littman97e}.
We have implemented this framework for decision making under
uncertainty in a language called Stochastic OPL. This is an
extension of the OPL constraint modelling language \cite{hent99}.
Finally, we describe a wide range of problems that we have modelled
in Stochastic OPL that illustrate some of its potential.

\section{Motivation Example \label{motivation}}

We consider a stochastic version of the
``template design'' problem.
The \textit{deterministic} template design problem (prob002 in
CSPLib, http://www.csplib.org) is described as follows. We are given a
set of variations of a design, with a common shape and size and such
that the number of required ``pressings'' of each variation is
known. The problem is to design a set of templates, with a common
capacity to which each must be filled, by assigning one or more
instances of a variation to each template. A design should be chosen
that minimises the total number of ``runs'' of the templates
required to satisfy the number of pressings required for each
variation. As an example, the variations might be for cartons for
different flavours of cat food, such as fish or chicken, where ten
thousand fish cartons and twenty thousand chicken cartons need to be
printed. The problem would then be to design a set of templates by
assigning a number of fish and/or chicken designs to each template
such that a minimal number of runs of the templates is required to
print all thirty thousand cartons. Proll and Smith address this
problem by fixing the number of templates and minimising the total
number of pressings \cite{ps98}.

In the \textit{stochastic} version of the problem, the demand for each
variation is uncertain. We adopt the Proll--Smith model in what follows,
extending it to comply with the stochastic demand assumption. We use
the following notation for problem parameters: $N$, number of
variations; $T$, number of templates; $S$, number of slots on each
template; $c_h$, unit scrap cost for excess inventory; $c_p$, unit shortage cost. The
decision variables are: $a_{i,j}$, number of slots designated to
variation $i$, on template $j$; $R_j$, number of required ``runs''
of template $j$. For convenience, we define the following auxiliary
variables: $x_i$, total production for variation $i$; $e_i$, total
scrap in variation $i$; $b_i$, total shortage in variation $i$.
There are also stochastic variables $d_i$ representing stochastic
demand for variation $i$.

This problem can be modelled as stochastic constraint
optimization problem. There is a
constraint to ensure that the total number of slots designated to
variations is exactly the number of slots available, which is $S$.
\begin{equation}
\sum_{i = 1}^N a_{ij} = S, \hspace{1em} \forall j \in \{1,...,T\},
\end{equation}
There is also a constraint to determine the total production in each
variation.
\begin{equation}
\sum_{j = 1}^T a_{ij}R_j = x_i, \hspace{1em} \forall i \in
\{1,...,N\}.
\end{equation}
And there are two constraints to determine the amount of shortage
and scrap for each variation.
\begin{eqnarray}
& &e_i = \max \{0,x_i - d_i\}, \forall i \in \{1,...,N\} \label{eq3}\\
& &b_i = -\min \{0,x_i - d_i\}, \forall i \in \{1,...,N\} \label{eq4}.
\end{eqnarray}

Our objective is to minimise the total
expected shortage and scrap costs,
\begin{equation}
\min E\left(\sum_{i = 1}^N c_p b_i + c_h e_i\right)
\end{equation}
where $E(.)$ denotes the expectation operator. From Eqs. (\ref{eq3}) and (\ref{eq4})
it is clear that $b_i$ and $e_i$ are random variables, since their values depend
on the realization of random demand.

Demand uncertainty necessitates carrying buffer-stocks. Overstock
leads to high inventory holding and scrap costs. On the other hand,
insufficient buffer stocks are also financially damaging, leading
to stock-outs and loss of customer satisfaction.
An alternative method to deal with such
uncertainty is to
introduce service-level constraints instead of using shortage costs.
In this case a service-level
constraint is expressed in the form of a chance constraint as
follows,
\begin{equation}
\mbox{Pr} \{ x_i - d_i \geq 0 \} \geq \alpha_i, \hspace{1em} i \in \{1,...,N\}
\end{equation}
where $\mbox{Pr}(.)$ represents the probability function and
$\alpha$ denotes a target service-level.

It may also be natural to consider measures such as the worst case
performance, other moments of expected performance like variance
which is a proxy for risk, the probability of attaining a
predetermined performance goal, and even, in certain types of
problems like engineering design problem, we may want to minimize
the spread (i.e. minimize the difference between the least and the
largest value of the objective function).
Note that stochastic variables need not be
independent (as assumed in \cite{walsh02}).
For example, if demand for a certain item is low in the first quarter,
it is more likely to be low in the second quarter as well.

Unfortunately, each of these extensions requires a major
modification to the backtracking and forward checking algorithms
presented in \cite{walsh02}. We therefore take a different track
which permits us to define these extensions without major
modifications to the solution methods. We define a new and
equivalent semantics for stochastic constraint programs
based on scenarios which permits the above
extensions, namely conditional probabilities, multiple chance
constraints, as well as a much wider range of goals. This
scenario-based view permits stochastic constraint programs to be
compiled down into regular (non-stochastic) constraint programs. We
can therefore use traditional constraint satisfaction and
optimization algorithms, as well as hybrid methods that use
techniques like integer linear programming.

\section{Scenario-based semantics}

A stochastic constraint satisfaction problem consists of a 6-tuple
$\langle V, S, D, P, C, \theta\rangle$.  $V$ is a set of decision
variables, and $S$ is a set of stochastic variables. $D$ is a
function mapping each element of $V$ and each element of $S$ to a
domain of potential values. A decision variable in $V$ is {\em
assigned} a value from its domain. $P$ is a function mapping each
element of $S$ to a probability distribution for its associated
domain. $C$ is a set of constraints, where a constraint $c\in C$ on
variables $x_i, \ldots, x_j$ specifies a subset of the Cartesian
product $D(x_i)\times\ldots\times D(x_j)$ indicating
mutually-compatible variable assignments. The subset of $C$ that
constrain at least one variable in $S$ are {\em chance constraints},
$h$. $\theta_h$ is a threshold probability in the interval $[0,1]$,
indicating the fraction of scenarios in which the chance constraint
$h$ must be satisfied. Note that a chance constraint with a
threshold of 1 is equivalent to a hard constraint.

A stochastic CSP consists of a number of {\em decision stages}. In a
one-stage stochastic CSP, the decision variables are set before the
stochastic variables. In an $m$-stage stochastic CSP, $V$ and $S$
are partitioned into $n$ disjoint sets, $V_1, \ldots,V_m$ and $S_1,
\ldots,S_m$. To solve an $m$-stage stochastic CSP an assignment to
the variables in $V_1$ must be found such that, given random values
for $S_1$, assignments can be found for $V_2$ such that, given
random values for $S_2$, $\ldots$, assignments can be found for
$V_m$ so that, given random values for $S_m$, the hard constraints
are satisfied and the chance constraints are satisfied in the
specified fraction of all possible scenarios.

In the policy based view of stochastic constraint programs of
\cite{walsh02}, the semantics is based on a tree of decisions. Each
path in a policy represents a different possible scenario (set of
values for the stochastic variables), and the values assigned to
decision variables in this scenario. To find satisfying policies,
backtracking and forward checking algorithms, which explores the
implicit AND/OR graph, are presented. Stochastic variables give AND
nodes as we must find a policy that satisfies all their values,
whilst decision variables give OR nodes as we only need find one
satisfying value.
An alternative semantics for stochastic constraint programs, which
suggests an alternative solution method, comes from a scenario-based
view \cite{birge97}.

\begin{figure}[htbp]
\centering \epsfig{file=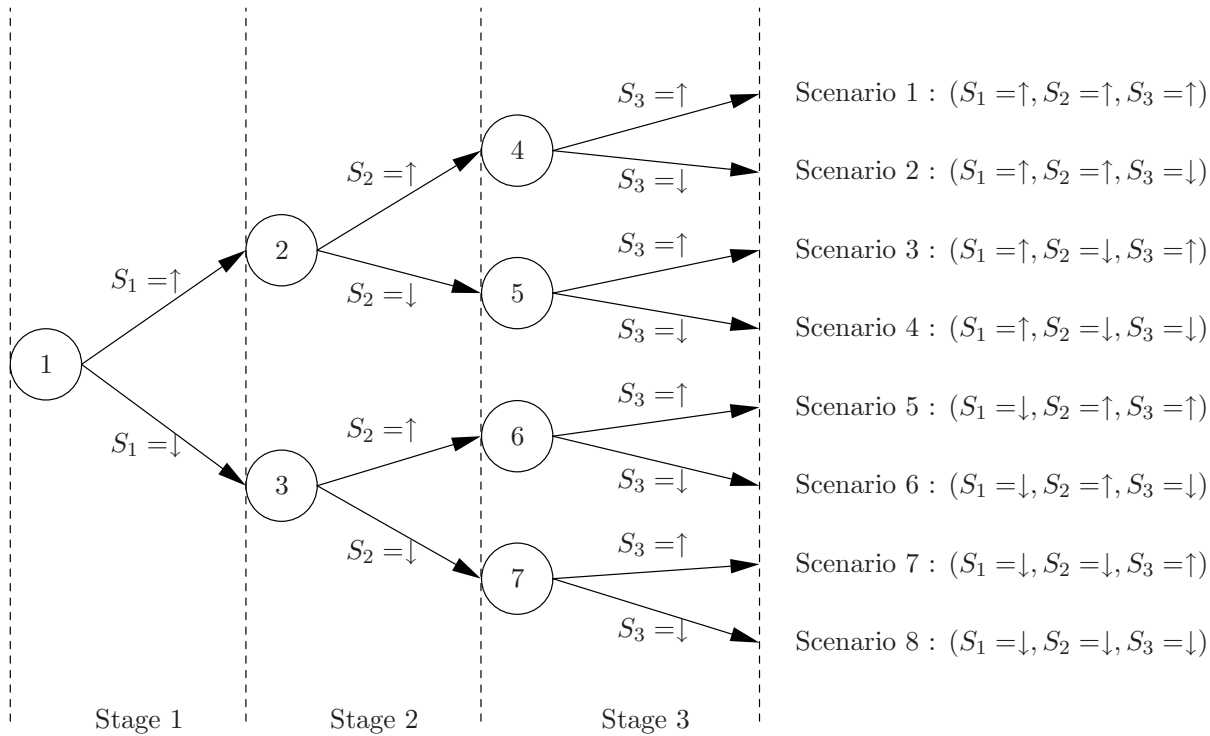, width=16cm}
\caption{A tree
representation of a Stochastic CP}\label{tree}
\end{figure}

In the scenario-based approach, a scenario tree is generated which
incorporates all possible realisations of discrete random
variables into the model explicitly. A tree representation of a
3-stage problem, with 2 possible states at each stage, is given in
Fig.\ref{tree}.


Scenarios deal with uncertain aspects (e.g. the economic
conditions, the state of the financial markets, the level of
demand) of the operating environment relevant to the problem.
Hence, the future uncertainty is described by a set of alternative
scenarios. The number of scenarios as well as the progression of
the scenarios from one period to another is problem specific. A
path from the root to an extremity of the event tree represents a
\textit{scenario}, $\omega \in \Omega$, where $\Omega$ is the set
of all possible scenarios. With each scenario a given probability
is associated. If $S_i$ is the $i$th random variable on a path
from the root to the leaf representing scenario $\omega$,
and $a_i$ is the value given to $S_i$ on the $i$th stage of
this scenario, then the probability of this scenario is
again $\prod_{i} \text{Pr}(S_i = a_i)$.

Thus, a scenario is associated with each path in the policy.
Within each scenario, we have a conventional (non-stochastic)
constraint program to solve. We simply replace the stochastic
variables by the values taken in the scenario, and ensure that the
values found for the decision variables are consistent across
scenarios as certain decision variables are shared across
scenarios. For instance, node 1 of the tree in Figure \ref{tree}
corresponds to the first stage and associated decisions are
identical for all scenarios. Note that in stage 2, the decisions of
scenarios 1 to 4 are identical. Similarly in stage 2, the decisions of
scenarios 5 to 8 are identical.

Constraints are defined (as in traditional constraint
satisfaction) by relations of allowed tuples of values, and can be
implemented with specialized and efficient algorithms for
consistency checking. The great advantage of this approach is that
we can use conventional constraint solvers to solve stochastic
constraint programs. We do not need to implement specialized
solvers. The scenario-based view of stochastic constraint programs
also allows later stage stochastic variables to take values which
are conditioned by the earlier stage stochastic variables. This is
a direct consequence of employing the scenario representation, in
which stochastic variables are replaced with their scenario
dependent values.

Of course, there is a price to pay as the number of scenarios grows
exponentially with the number of stages. However, our results show
that a scenario-based approach is feasible for many problems.
Indeed, we observe much better performance using scenario-based
approach on the book production planning example of Walsh
\cite{walsh02} compared to the tree search methods. In addition, as
we discuss later, we have developed a number of techniques like
Latin hypercube sampling to reduce the number of scenarios
considered.

\begin{table*}[h]
\centering
\begin{tabular}{|c|rr|rr|rrr|}\hline
& \multicolumn{2}{c|}{Backtracking (BT)} &
\multicolumn{2}{c|}{Forward Checking (FC)} &
\multicolumn{3}{c|}{Scenario-Based (SB)} \\ No. Stages &
\multicolumn{1}{c}{Nodes} & CPU/sec & \multicolumn{1}{c}{Nodes} &
CPU/sec & Failure & Choice Points & CPU/sec
\\\hline
1 & 28 & 0.01 & 10 & 0.01 & 4 & 5 & 0.00  \\
2 & 650 & 0.09 & 148 & 0.03 & 4 & 8 & 0.02 \\
3 & 17,190 & 2.72 & 3,604 & 0.76 & 8 & 24 & 0.16 \\
4 & 510,356 & 83.81 & 95,570 & 19.07 & 42 & 125 & 1.53 \\
5 & 15,994,856 & 3,245.99 & 2,616,858 & 509.95 & 218 & 690 & 18.52 \\
6 & -- & -- & -- & -- & 1260 & 4035 & 474.47 \\\hline
\end{tabular}
\caption{A Comparison of Policy Search Methods and Scenario Tree
Approach \label{table1}}
\end{table*}

The results in Table \ref{table1} on the book production planning
example from \cite{walsh02} show that the scenario-based approach
offers much better performance on this problem than the forward
checking or backtracking tree search algorithms.
Failures and choice points denote the number of failures encountered
during the resolution and the number of choices needed to produce
the solution, respectively.


Constraint satisfaction is NP-complete in general. Not surprisingly,
stochastic constraint satisfaction moves us up the complexity
hierarchy.

\newtheorem{mytheorem}{Theorem}
\newcommand{\myproof}{\noindent {\bf Proof:\ \ }}
\newcommand{\myqed}{\mbox{$\diamond$}}

\newpage

\begin{mytheorem}
Stochastic constraint satisfaction problems are PSPACE-complete.
\end{mytheorem}
\myproof Membership in PSPACE follows from the existence of a naive
depth-first and/or search tree algorithm for solving stochastic
constraint satisfaction problems. This algorithm recurses through
the variables in order, making an ``and'' branch for a stochastic
variable and an ``or'' branch for a decision variable. The algorithm
requires linear space in the number of variables.

To show completeness, we reduce the satisfiability of quantified
Boolean formula to a stochastic constraint satisfaction problem.
Each existential Boolean variable in the quantified Boolean formula
is mapped to a Boolean decision variable in the stochastic CSP. Each
universal Boolean variable in the quantified Boolean formula is
mapped to a Boolean stochastic variable in the stochastic CSP. Such
variables have equal probability of being $true$ or $false$. Each
clause is replaced by the equivalent constraint which is to be
satisfied in all possible scenarios. The reduction is linear in the
size of the original quantified Boolean formula. The quantified
Boolean formula is satisfiable iff the stochastic CSP is itself
satisfiable. \myqed

\section{Stochastic OPL \label{sopl}}

We have implemented this framework on top of the OPL constraint
modelling language \cite{hent99}. An OPL model consists of two
parts: a set of declarations, followed by an instruction.
Declarations define the data types, constants and decision
variables. An OPL instruction is either to satisfy a set of
constraints or to maximize/minimize an objective function subject to
a set of constraints. We have extended the declarations to include
the declaration of stochastic variables, and the instructions to
include chance constraints, and a range of new goals like maximizing
the expectation of an objective function.

\subsection{Constant and Variable declarations}

Stochastic variables are set according to a probability
distribution using a command of the form:
\begin{verbatim}
 stoch <Type> <Id> <Dist>;
\end{verbatim}
Where {\tt <Type>} is (as with decision variables) a data type
(e.g. a range of values, or an enumerated list of values), {\tt
<Id>} is (as with decision variables) the variable name, and {\tt
<Dist>} defines the probability distribution of the stochastic
variable(s). Probability distributions include {\tt uniform}, {\tt
poisson(lambda)}, and user defined via a list of (not necessarily
normalized) values. Other types of distribution can be supported
as needed. We insist that stochastic variables are arrays, with
the last index describing the stage. Here are two different data
representations of stochastic variables:
\begin{verbatim}
 stoch float market[Years] =
     [<0.05 (0.34), 0.07 (0.66)>,<0.02 (0.25), 0.04 (0.25), 0.09 (0.5)>];
 stoch int demand[Period] =
     [2 (0.25), 3 (0.75), 4 (0.35), 5(0.15), 7 (0.50), 8 (0.40), 9 (0.60)];
\end{verbatim}
In the first, we have a float variable in the first year which is
either 0.05 (with probability 0.34) or 0.07 (with probability
0.66). The notation, ``$<.>$'', is convenient for problems in
which random variables are independent. In the second, we have the
stochastic variable \verb"demand", which takes the value of either
2 or 3 in the first period. The value of the random variable in
the second period depends on the first period's realization. It is
$\{4,5,7\}$ if the first period's demand is 2, and $\{8,9\}$ if it
is 3. This notation is convenient especially for problems
involving conditional probabilities.

The constants are declared as in OPL with the exception of the case
where their values depend on the stochastic variables. In a
financial planning problem, if it is assumed that the financial
instrument return rates solely depend on unpredictable market then
the return matrix \verb"return[Instr,Period]" must be related to the
stochastic variable \verb"market[Period]". The dependence of
\verb"return" on \verb"market" is denoted by joining them together
by a hat, $\hat{\hspace{0.5em}}$, as in

\verb"return[Instr,Period]^market = ...;"

\subsection{Constraint posting \label{const_posting}}

We can post both hard constraints (as in OPL) and chance
constraints. Chance constraints hold in some but not necessarily
all scenarios. They are posted using a command of the form:
\begin{verbatim}
 prob(<Constraint>) <ArithOp> <Expr>;
\end{verbatim}
Where {\tt <Constraint>} is any OPL constraint, {\tt <ArithOp>} is
any of the arithmetically comparison operations ({\tt =},{\tt
<>},{\tt <},{\tt >}, {\tt <=}, or {\tt >=}) and {\tt <Expr>} is
any arithmetic expression (it may contain decision variables or
may just be a rational or a float in the range 0 to 1). For
example, the following command specifies the chance constraint
that in each quarter the demand (a stochastic variable) does not
exceed the production (a decision variable) plus the stock carried
forward in each quarter (this auxiliary is modelled, as in
conventional constraint programming, by a decision variable) with
80\% probability:
\begin{verbatim}
 forall(i in 1..n)
    prob(demand[i] <= production[i]+stock[i]) >= 0.80;
\end{verbatim}
Constraints which are not chance constraints are hard and have to
hold in all possible scenarios. For example, the stock carried
forwards is computed via the hard constraint:
\begin{verbatim}
 forall(i in 1..n)
    stock[i+1] = max(0,stock[i] + production[i] - demand[i]);
\end{verbatim}

\subsection{Optimization \label{obj_func}}

Stochastic OPL supports both stochastic constraint satisfaction
and optimization problems. We can maximize or minimize the
expectation of an objective function.
%
As an example of an expected value function, we'll consider the
stochastic template design problem of Sec.\ref{motivation},
in which the expected total cost of scrap and shortage is minimised.
This can be specified by the following (partial) model:
\begin{verbatim}
 minimize expected(cost)
 subject to
   cost = sum(i in 1..n) ch*max(0,x[i]-d[i])-cp*min(0,x[i]-d[i]);
   ...
\end{verbatim}
We can also model risk. For example, we may wish
to reason about the mean and variance in the return
for a portfolio selection
problem \cite{markowitz52}.
Markowitz's mean/variance model provides a framework to examine
the tradeoff between the expected value and its variability. In
the mean/variance model, the expected value plus a constant
($\lambda\leq 0$) times the standard deviation --standard
deviation is used as a surrogate for risk-- is maximized. However,
since the risk expression of Markowitz is very complex to reason with,
we consider the simplification introduced
in \cite{konno91} and \cite{konno90}.

In \cite{konno90}, the absolute deviation function, $K$, is
introduced as $K = \left| Q - E\{Q\} \right|$, where the random
variable Q denotes the objective function value. \cite{konno91}
demonstrates that mean absolute deviation function can remove most
of the difficulties associated with the standard deviation
function.
Stochastic OPL supports Markowitz's mean/variance model, where the
surrogate risk measure is the absolute deviation function, with
the command:

\verb" maximize mv(<Expr>,"$\lambda$\verb")"

The mean absolute deviation risk model is discussed in
Sec.\ref{robust}.

Stochastic OPL also supports a number of other optimization goals.
For example:
\begin{verbatim}
 minimize spread(profit)
 maximize downside(profit)
 minimize upside(cost)
\end{verbatim}
The spread is the difference between the value of the objective
function in the best and worst scenarios, whilst the downside
(upside) is the minimum (maximum) objective function value a
possible scenario may take.

\section{Compilation of Stochastic OPL}

These stochastic extensions are compiled down into conventional
(non-stochastic) OPL models automatically by exploiting the
scenario-based semantics. The compiler is written in Lex and Yacc,
with a graphical interface in Visual C++. A demonstration version
of the stochastic OPL compiler and example problems can be downloaded
from http://www.cs.york.ac.uk/\ensuremath{\sim}at/project.
In what follows, the
compilation of decision variables, constants, hard constraints,
chance constraints and objective functions are discussed.

\subsection{Compiling Constants, Variables and Hard Constraints \label{hard_const}}

Compilation involves replacing stochastic variables by their
possible values, and decision variables by a ragged array of
decision variables, one for each possible scenario. Constants that
depend on stochastic variables also require ragged arrays. We
identify positions in the scenario tree by the stage (e.g. first
stage, or second stage) and by an ordering over the states at this
stage. We therefore declare:
\par\noindent \verb"struct AStruct {"
\\ \verb"       AStageRange stage;"
\\ \verb"       int state; };"
\\where \verb"AStageRange" is a stage index range and is extracted from the stochastic
variable declaration. By means of this structure, the relevant
$<$stage, state$>$ pairs are declared:
\par\noindent
\verb"{AStruct}"
\\ \verb"Nodes = { <stage,state> | stage in AStageRange & state in 1..nbNodes[stage] };"
\\where \verb"nbNodes[stage]" array denotes the number of states at the beginning of each stage and is extracted from probability data.

To build the certainty equivalent model using the notion of
scenarios, a matrix \verb"ScenTree" is declared and a reference to
each node is made via $<$\verb"stage,ScenTree[stage,scen]"$>$
where \verb"scen" denotes a scenario.
Variable and constant compilations are performed by means of
$<$\verb"stage,ScenTree[stage,scen]"$>$ notation and the following
rules:

\begin{itemize}

\item \textit{Constants with No-Stochastic Dependence}:
\\Constants declared as independent of stochastic
variable, are not altered.

\item \label{ref1} \textit{Constants with Stochastic Dependence}:
\\In the compiled model, in each stage for each random realization,
there should be a unique value of each constant with stochastic
dependence. This is achieved by replacing constants
\verb"C[...,t,...]^R" with \verb"C[...,<t,ScenTree[t+1,scen]>,...]"
where \verb"t" stands for the stage index.

\item \textit{Variables with No-Stage Index} :
\\ For each scenario there should be a unique decision variable. Therefore, a decision variable without a
stage index, \verb"X["$ind_1$\verb",...,"$ind_n$\verb"]", is
replaced with \verb"X["$ind_1$\verb",...,"$ind_n$\verb",scen]".

\item \label{end_item} \textit{Variables with Stage Index} :
\\Stage indexed variables  are modified the
same way as constants that depend on the stochastic variable:
\verb"X[...,<t,ScenTree[t,scen]>,...]" replaces \verb"X[...,t,...]".
\end{itemize}
Once the variables and constants are transformed and the range of
possible scenarios, \verb"Scenarios", is determined then the
compilation of stochastic hard constraints into equivalent
deterministic ones requires only a \verb"forall" statement to cover
all possible scenarios: \verb" forall(scen in Scenarios)" \{
\textit{all constraints} \};

The following financial planning example (see section
\ref{portfolio} for the problem description), with a stage range
\verb"1..N", demonstrates the application of compilation rules. A
mathematical formulation of the problem, and corresponding
stochastic and certainty-equivalent OPL representations thereof are
given in the appendix. As explained above, all problem constraints
must be given as a part of the \verb"forall(scen" \verb"in"
\verb"Scenarios)" \verb"{ <Constraints> };" statement.

\begin{itemize}

\item \verb"capital = wealth[1];" \textit{// ``capital'' is a
constant, ``wealth'' is a decision variable}
\\ is compiled into \verb"capital = wealth[<1,ScenTree[1,scen]>];"

\item \verb"finalwealth = wealth[N+1];" \textit{// ``finalwealth''
is a decision variable}
\\ is compiled into \verb"finalwealth[scen] = wealth[<N+1,ScenTree[N+1,scen]>];"

\item \verb"wealth[p] = sum(i in Instr) investment[i,p];"
\textit{// ``investment'' is a decision variable}
\\ is compiled into \verb"wealth[<p,ScenTree[p,scen]>] ="
\\ \verb"               sum(i in Instr) investment[i,<p,ScenTree[p,scen]>];"

\item \verb"wealth[p+1] = sum (i in Instr)"
\\\verb"        investment[i,p]*(1+ret[i,p]^market);"
\\ is compiled into
\\ \verb"  wealth[<p+1,ScenTree[p+1,scen]>] = sum (i in Instr)"
\\ \verb"      investment[i,<p,ScenTree[p,scen]>]*(1+ret[i,<p,ScenTree[p+1,scen]>]);"

\end{itemize}

\subsection{Compiling Chance Constraints}

The chance constraints posted using a command of the form
\begin{verbatim}
 prob(<Constraint>) <ArithOp> <Expr>;
\end{verbatim}
are compiled into a sum constraint of the form
\begin{verbatim}
 sum(scen in Scenarios)
    Probability[scen]*(<Constraint[scen]>) <ArithOp> <Expr>;
\end{verbatim}
where {\tt <Constraint[scen]>} is a compiled Stochastic OPL
constraint in scenario {\tt scen}, {\tt Probability[scen]} the
probability of scenario {\tt scen} and \verb+Scenarios+ the set of
all scenarios.

As an example of chance constraint compilation consider the
following inventory constraints:
\begin{verbatim}
forall (p in Periods) Stock[p]+Order[p]-Demand[p] = Stock[p+1];
forall (p in Periods) Stock[p+1] >= 0;
\end{verbatim}
These are inventory balance equations and non-negative stock
constraints, respectively. In the case of stochastic demand, it is
generally very expensive to follow a policy which guarantees no
backlogging, i.e., meeting all customers' demand. Instead,
generally, a target service level is introduced by the management and the
complete demand satisfaction policy is relaxed. Hence, the
inventory problem now becomes:
\begin{verbatim}
forall (p in Periods) Stock[p]+Order[p]-Demand[p] = Stock[p+1];
forall (p in Periods) prob(Stock[p+1] >= 0) >= ServLev;
\end{verbatim}
The inventory balance equation, which is a hard constraint, can be
compiled into its certainty equivalent form as explained in
Sec.\ref{hard_const}. The compilation of chance constraints are
done in a similar fashion, with the only exception of the
introduction of weights, which are actually the probabilities of
relevant scenarios. The service-level expression is compiled into
the following OPL constraint:
\begin{verbatim}
forall (p in Periods)
  sum (scen in 1..nbNodes[p+1])
  (Stock[<p+1,ScenTree[p+1,scen]>] >= 0)*Probability[<p+1,scen>] >= ServLev;
\end{verbatim}

Note that the bracketing of the inequality reifies the constraint
so that it takes the value 1 if satisfied and 0 otherwise.

\subsection{Compiling Objective Functions \label{tranf_obj}}

The most common objective function type, the expected value
function, is incorporated into Stochastic OPL with the reserved
word \verb"expected" and can be compiled into standard OPL the
same way as \verb"prob":
\begin{verbatim}
maximize expected(<Expr>)
\end{verbatim}
compiles into
\begin{verbatim}
maximize
   sum(scen in Scenarios)
      (<Expr[scen]>)*Probability[scen]
\end{verbatim}

Another important objective function type, Markowitz's
mean/variance model, \verb"mv(<Expr>,"$\lambda$\verb")" or
$E\{Q\}+\lambda E\{K\}$, which takes into account the tradeoff
between the expected value of an objective function and its
variability, can easily be expressed in the certainty equivalent
form by employing the absolute deviation function of
Sec.\ref{obj_func}:
\\\verb"        minimize"
\\\verb"          expected(Q) + "$\lambda$\verb"*expected(K)"
\\\verb"        subject to {"
\\\verb"           K - Q + expected(Q) "$\geq$\verb" 0;"
\\\verb"           K + Q - expected(Q) "$\geq$\verb" 0;"
\\\verb"           ... };"

Other objective functions, namely,
\begin{verbatim}
 minimize spread(profit)
 maximize downside(profit)
 minimize upside(cost)
\end{verbatim}
which are supported by Stochastic OPL are compiled in a similar
fashion.

This is by no means an exhaustive list but it gives an indication
of the variety of optimization goals and the versatility of the
Stochastic OPL system. New optimization goals can be incorporated
into the system as needed.

\section{Value of information and stochastic solutions \label{evpi}}

We can also easily provide the user with information about how
much value is obtained if we were to know the value
of stochastic variables. For example, in some situations it can be
possible to wait for stochastic variables to realize their values.
Alternatively, we can show how expensive it is to fix
to a solution now that ignores future changes.

Consider a payoff table based on $m$ possible decisions $D_i$ for
$i=1,...,m$ and $n$ possible scenarios $S_j$ for $j=1,...,n$. The
payoff for decision $D_i$, if scenario $S_j$ will occur, is
$a_{ij}$. Suppose the probability that scenario $S_j$ will occur
is $p_j$. If the decision criterion is the expected payoff, then
the best decision is the one that maximizes $\sum_{j=1}^{n} p_j
a_{ij}$ and the solution is called ``stochastic solution'', SS.

There is a family of models, one for each scenario, and the
weighted average of solutions for each scenario (solved by
assuming that all data were already known) gives the expected
``wait-and-see solution'', WSS.

Consider the hypothetical situation that one knows ahead of time
which scenario will occur. If such information is available then
one may expect extra payoff with a non-negative value. The
expected value of payoff can serve as an upper bound for the value
of information, which is called the ``expected value of perfect
nformation'' (EVPI). It is assumed that the probability that the
perfect information will indicate that scenario $S_j$ will occur
is also $p_j$.

From the definition, the EVPI is the difference between the
expected payoff calculated using the maximum $a_{ij}$ for each
scenario (WSS) and the expected payoff for the best decision (SS).

The expression for the EVPI is thus:
\begin{equation}\nonumber
EVPI = WSS - SS =  \sum_{j=1}^{n} p_j \max_{\substack{1 \leq i
\leq m}} \{a_{ij}\} - \max_{\substack{1 \leq i \leq m}}
\left\{\sum_{j=1}^{n} p_j a_{ij}\right\}
\end{equation}

This is therefore the most that should be spent in gathering
information about the uncertain world.

For stochastic optimization problems, we may compute another
statistics which quantify the importance of randomness. The
``value of stochastic solution'', VSS, measures the possible gain
from solving the stochastic model that explicitly incorporates the
distribution of random variables within the problem formulation.

Some models do not take into account the randomness of different
uncertain parameters. They replace the uncertain parameters by
their expected values and solve then the so called expected value
problem. It means that only one scenario, namely the expected
value scenario, is considered. In this case the solution to the
expected value scenario will give an objective function value for
the stochastic problem, which is called the ``expected value
solution'', EVS.

The value of a stochastic solution (VSS) is then the difference
between SS and EVS:
\begin{eqnarray*}
VSS  =  SS - EVS \geq 0.
\end{eqnarray*}
This computes the benefit of knowing the distributions of the
stochastic variables. It is a well-known fact in decision theory
that the above relation is valid. This means that the objective
function's expected value of the stochastic optimization problem
will be better than the expected value of deterministic
programming.

These statistics can easily be calculated using our framework.
Such calculation requires the solution of $n$ independent
deterministic scenario problems to determine WSS and a single
expected value problem to determine EVS.

\section{Scenario reduction}

Each scenario introduces new decision variables. For many
practical problems, it is too expensive to compute all possible
scenarios. How then can we replace a large, computationally
intractable scenario tree with a small, tractable tree so that
solving the problem over the small tree yields a solution not much
different than the solution over the large tree?

We have implemented several techniques to reduce the number of
scenarios. These scenario reduction algorithms determine a subset
of scenarios and a redistribution of probabilities relative to the
preserved scenarios. No requirements on the stochastic data
process are imposed and therefore the concept is general. However,
the reduction algorithms, depending on their sophistication, may
require different types of data.

The simplest scenario reduction algorithm is to consider just a
single scenario in which stochastic variables take their expected
values. This is supported with the command:
\begin{verbatim}
 scenario expected;
\end{verbatim}
This is actually the aforementioned (see Sec.\ref{evpi}) expected
value problem. Among the other methods presented here, this is the
most crude one.

The user may also be content to consider just a few of
the most probable
scenarios and ignore rare events. This method is referred as
``mostlikely'' in the rest of the paper. We support this with the
command:
\begin{verbatim}
 scenario top <Num>;
\end{verbatim}
Another option is to use Monte Carlo sampling. The user can
specify the number of scenarios to sample using a command of the
form:
\begin{verbatim}
 scenario sample <Num>;
\end{verbatim}
The probability distributions of the stochastic variables is used
to bias the construction of these scenarios.

We can also consider sampling methods which may
converge faster than simple Monte Carlo sampling.
For example, we implemented one of the best sampling methods from
experimental design, and one of the best scenario reduction
methods from operations research. Latin hypercube sampling (LHS)
\cite{mckay79}, ensures that a range of values for a variable are
sampled. Suppose we want the sample size to be $n$. We divide the
unit interval into $n$ intervals, and sample a value for each
stochastic variable whose cumulative probability occurs in each of
these interval. We then construct $n$ sample scenarios from these
values, enforcing the condition that the samples use each value
for each stochastic variable exactly once. More precisely, let
$f_i(a)$ be the cumulative probability that $X_i$ takes the value
$a$ or less, $P_i(j)$ be the $j$th element of a random permutation
$P_i$ of the integers $\{0,\ldots,n-1\}$, and $r$ be a random
number uniformly drawn from $[0,1]$. Then, the $j$th Latin
hypercube sample value for the stochastic variable $X_i$ is:
$$ f_i^{-1}(\frac{ P_i(j) + r}{n}) $$
However, it should be noted that the sample size $n$ does not
guarantee to produce a sample of $n$ scenarios, since a single
scenario may be chosen more than once due to, for example,
the discreteness of the data. The command for LHS is
\begin{verbatim}
 scenario lhs <Num>;
\end{verbatim}
where \verb"<Num>" denotes the number of non-overlapping intervals
used with LHS.

Finally, we implemented a scenario reduction method used in
stochastic programming due to Dupacova, Growe-Kuska and Romisch
\cite{dupacova03}. Dupacova et al. assume that the original
probability measure $P$ is discrete with finitely many scenarios
and this probability measure is approximated by a probability
measure $Q$ of a smaller number of scenarios. In this
case, the upper bound for the distance between $P$ and $Q$ is a
Kantorovich functional. Then the upper bound represents the
optimal value of a Monge-Kantorovich mass transportation problem.
The Monge-Kantorovich mass transportation problem dates back to
work by Monge in 1781 on how to optimally move material from one
place to another, knowing only its initial and final spatial
distributions, the cost being a prescribed function of the distance
travelled by molecules of material \cite{rachev98a}, \cite{rachev98b}.
Dupacova et al. show that the Kantorovich functional of the
original probability distribution $P$ and the optimal reduced
measure $Q$ based on a given subset of scenarios of $P$ as well as
the optimal weights of $Q$ can be computed without solving
the Monge-Kantorovich problem.
Their backward reduction algorithm for
determining a subset of scenarios is given below.

Let $n_T$ denote the number of stages of the optimization problem
and $n_s$ the number of scenarios. A scenario $\omega^{(i)}$, $i
\in \{1,...,n_s\}$, is defined as a sequence of nodes of the tree
\begin{equation}\nonumber
    \omega^{(i)} = ( \eta_0^{(i)}, \eta_1^{(i)},...,\eta_{n_T}^{(i)}
    ),\hspace{2em}    i=1,...,n_s
\end{equation}
For each node belong to scenario $j$ on stage $s$, a vector
$\mathrm{\mathbf{p}}_s^{(j)} \in \mathbb{R}^{n_s^p}$ of parameters
is given. The probability to get from $\eta_j^{(i)}$ to
$\eta_{j+1}^{(i)}$ is denoted by $\pi_{j,j+1}^{(i)}$. Thus the
probability for the whole scenario $\omega^{(i)}$ is given by
\begin{equation}\nonumber
    \pi^{(i)} = \prod\limits_{j=0}^{n_T -1} \pi_{j,j+1}^{(i)}
\end{equation}
The distance between two scenarios $\omega^{(i)}$ and
$\omega^{(j)}$ is defined as
\begin{equation}\nonumber
    d(\omega^{(i)},\omega^{(j)})=\sqrt{\sum\limits_{s=0}^{n_T}(\mathrm{\mathbf{p}}_s^{(i)}-\mathrm{\mathbf{p}}_s^{(j)})^2}
\end{equation}
according to the Euclidean norm in the space of the parameter
vectors.

The scenario deletion procedure given below is applied iteratively,
deleting one scenario in each iteration, until a given number of
scenarios remains.

\begin{itemize}

\item[S1.] Determine the scenario to be deleted: Remove scenario
$\omega^{(s^*)}$, $s^* \in \{1,...,n_s\}$ satisfying
\begin{equation}\nonumber
    \pi^{(s^*)} \min\limits_{\bar{s} \neq s^*} d(\omega^{(\bar{s})},\omega^{(s^*)}) =
    \min\limits_{m,n \in \{1,...,n_s\}} \{ \pi^{(m)} \min\limits_{n \neq m}
    d(\omega^{(n)},\omega^{(m)})\}
\end{equation}
Hence, not only the distances, but also the probabilities of the
scenarios are considered.

\item[S2.] Change the number of scenarios: $n_s := n_s - 1$.

\item[S3.] Change the probability of the scenario
$\omega^{(\bar{s})}$, that is the nearest to $\omega^{(s^*)}$:
\begin{equation}\nonumber
    \pi^{(\bar{s})} := \pi^{(\bar{s})}+\pi^{(s^*)}
\end{equation}
\begin{equation}\nonumber
    \pi^{(s^*)} := 0
\end{equation}
All node probabilities are adjusted, as the sum of the
probabilities of possible realizations at each node equals 1.

\item[S4.] If $n_s>N$ then go to step 1, where $N$ is the desired
number of scenarios remain.

\end{itemize}

This algorithm is incorporated into Stochastic OPL and can be
called by
\begin{verbatim}
 scenario DGR <Num>;
\end{verbatim}

Dupacova et al. report power production planning problems on which
this method offers 90\% accuracy sampling 50\% of the scenarios
and 50\% accuracy sampling just 2\% of the scenarios.

\section{Some examples \label{examples}}

To illustrate the potential of this framework for decision making
under uncertainty, we now describe a wide range of problems that
we have modelled. In each problem, we illustrate the effectiveness
of different scenario reduction techniques.

\subsection{Portfolio Diversification \label{portfolio}}

The portfolio diversification problem of \cite{birge97} can be
modelled as a stochastic COP. Suppose we have $\$P$ to invest in any
of $I$ investments and we wish to exceed a wealth of $\$G$ after $t$
investment periods. To calculate the utility, we suppose that
exceeding $\$G$ is equivalent to an income of $q\%$ of the excess
while not meeting the goal is equivalent to borrowing at a cost
$r\%$ of the amount short. This defines a concave utility function
for $r>q$. The uncertainty in this problem is the rate of return,
which is a random variable, on each investment in each period. The
objective is to determine the optimal investment strategy, which
maximizes the investor's expected utility. A mathematical
formulation of the problem, and corresponding stochastic and
certainty-equivalent OPL representations are given in the
appendix.

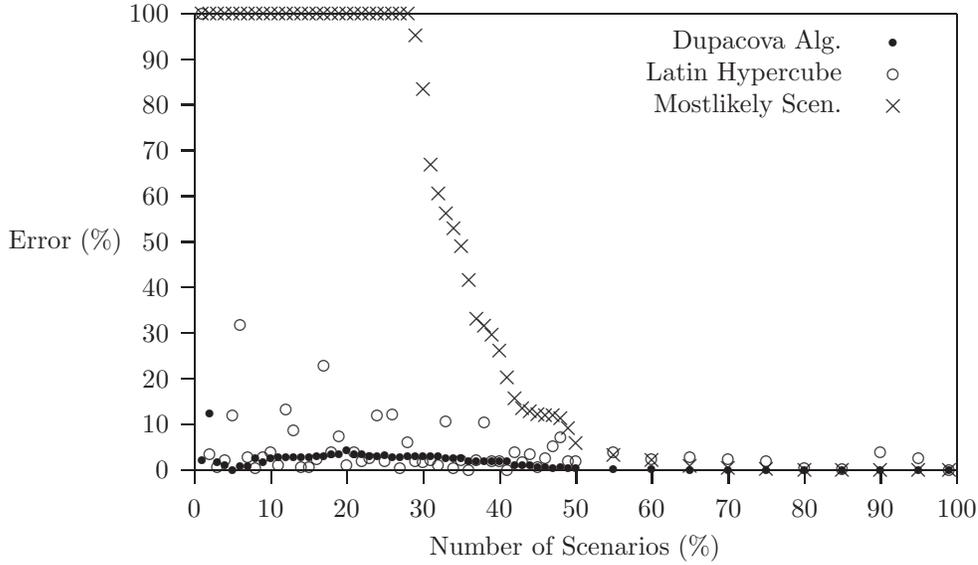
\begin{figure}[h]
\begin{center}
\input{sb1.tex}
\end{center}
\caption{Portfolio Diversification -- I} \label{fig1}
\end{figure}

The test problem has 8 stages, in which the number of states are
[5,4,4,3,3,2,2,2], and 5760 scenarios. The CP model has 27 decision
variables and 18 constraints for one scenario, and 33,438 decision
variables and 22,292 constraints for 5,760 scenarios. To compare the
effectiveness of different scenario reduction algorithms, we
adopt a two step procedure. In the first step, the scenario reduced
problem is solved and the first period's decision is observed. We
then solve the full-size (non scenario reduced) problem to
optimality with this first decision fixed. The difference between
the objective values of these two solutions is normalized by the
range [optimal solution, observed worst solution] to give a
normalized error for committing to the scenario reduced first
decision. In Fig. \ref{fig1}, we see that Dupacova et al's algorithm
is very effective, that Latin hypercube sampling is a small distance
behind, and both are far ahead of the most likely scenario method
(which requires approximately half the scenarios before the first
decision is made correctly).

\begin{figure}[h]
\begin{center}
\input{sb2.tex}
\end{center}
\caption{Portfolio Diversification -- II} \label{fig2}
\end{figure}
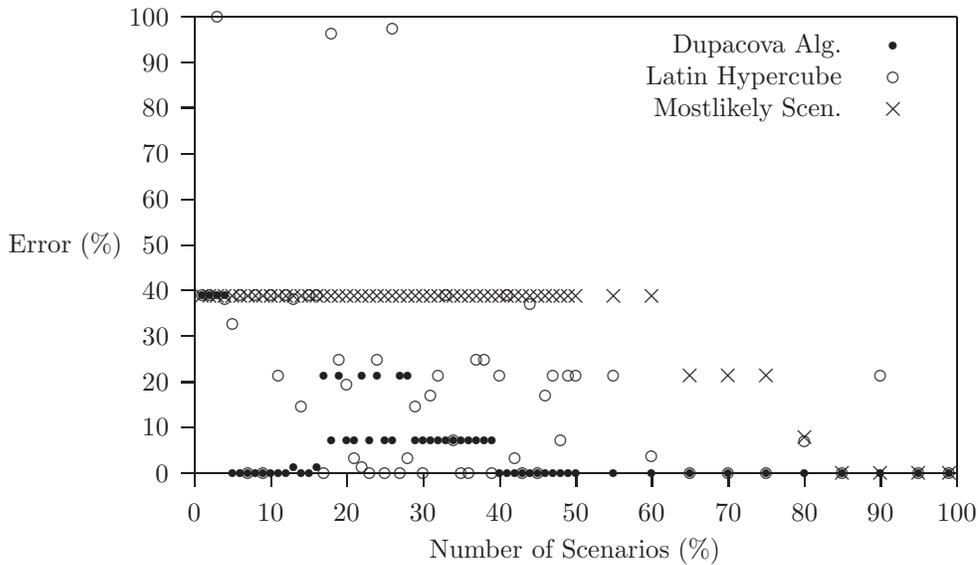

A smaller version of the above problem is designed with 4 stages,
[5,4,4,3] states in each stage and hence 240 scenarios in total. The
CP model has 15 decision variables and 10 constraints for one
scenario, and 1,038 decision variables and 692 constraints for 240
scenarios. Fig.\ref{fig2} shows that Dupacova et al. algorithm
requires approximately one third the scenarios before the first
decision is made correctly. It is interesting to see that the
general performance of scenario reduction algorithms has
deteriorated in the 4-stage case. This is mainly due to the fact
that the longer the planning horizon, the better the chance of recovery
from an early mistake.

\subsection{Agricultural Planning}

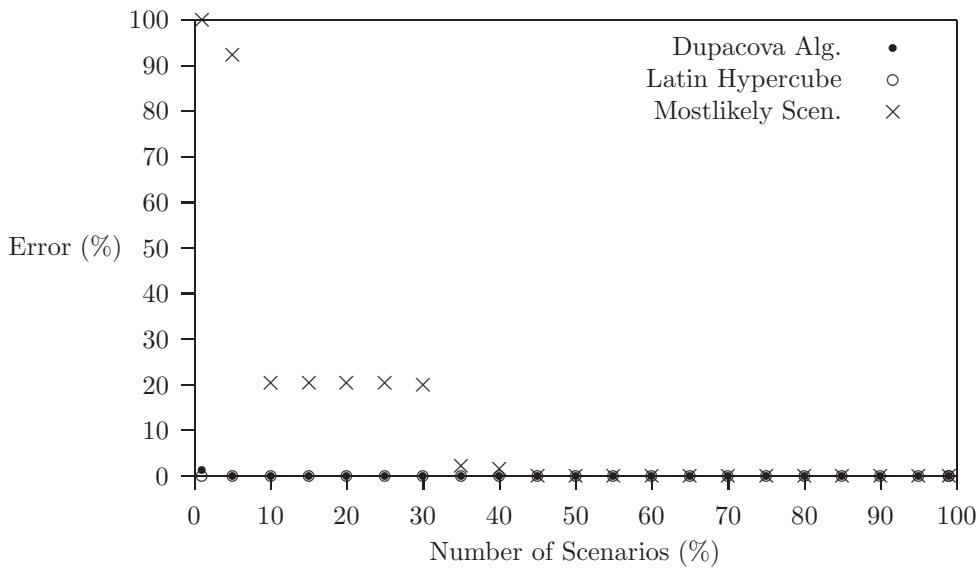
\begin{figure}[h]
\begin{center}
\input{farm1.tex}
\end{center}
\caption{Agricultural Planning -- I} \label{fig3}
\end{figure}

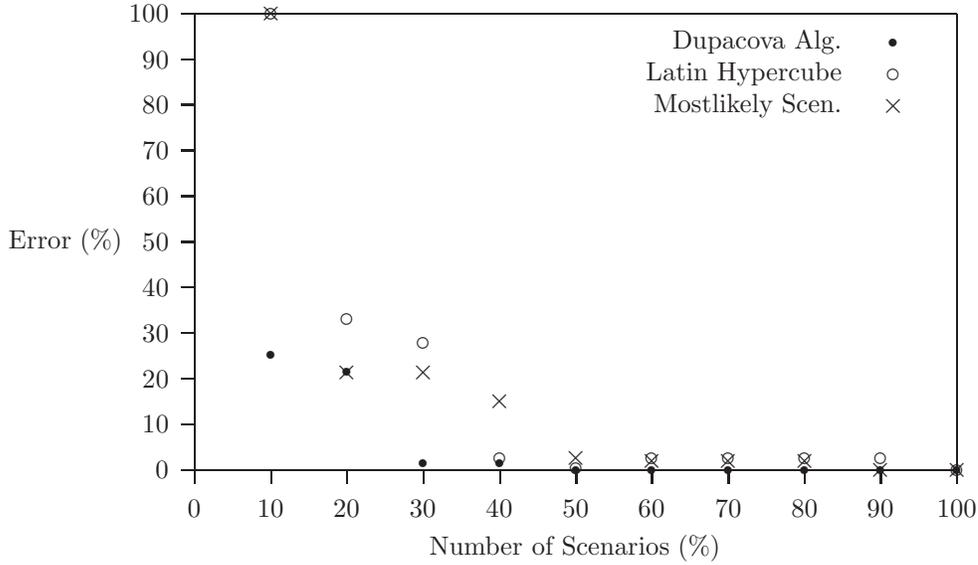
\begin{figure}[h]
\begin{center}
\input{farm3.tex}
\end{center}
\caption{Agricultural Planning -- II} \label{fig5}
\end{figure}

Farmers must deal with uncertainty since weather and many other
factors affect crop yields. In this example (also taken from
\cite{birge97}), we must decide on how many acres of his fields to
devote to various crops before the planting season. A certain amount
of each crop is required for cattle feed, which can be purchased
from a wholesaler if not raised on the farm. Any crop in excess of
cattle feed can be sold up to the EU quota; any amount in excess of
this quota will be sold at a low price. Crop yields are uncertain,
depending upon weather conditions during the growing season. This
test problem (Agricultural Planning -- I) has 4 stages and 10,000
scenarios. The CP model has 55 decision variables and 30 constraints
for one scenario, and 163,324 decision variables and 116,661
constraints for 10,000 scenarios. In Fig. \ref{fig3}, we again see
that Dupacova et al's algorithm and Latin hypercube sampling are
very effective, and both are far ahead of the most likely scenario
method. Fig.\ref{fig5} shows the results for a smaller instance
(Agricultural Planning -- II), with 1 stage and 10 scenarios only.
The CP model has 16 decision variables and 9 constraints for one
scenario, and 124 decision variables and 81 constraints for 10
scenarios. The adverse effect of the shorter planning horizon on the
effectiveness of scenario reduction techniques is also observed in
this case.

\subsection{Production/Inventory Management \label{proinv}}

\begin{figure}[h]
\begin{center}
\input{t_s.tex}
\end{center}
\caption{Production/Inventory Management (Full Cost) }
\label{fig6}
\end{figure}
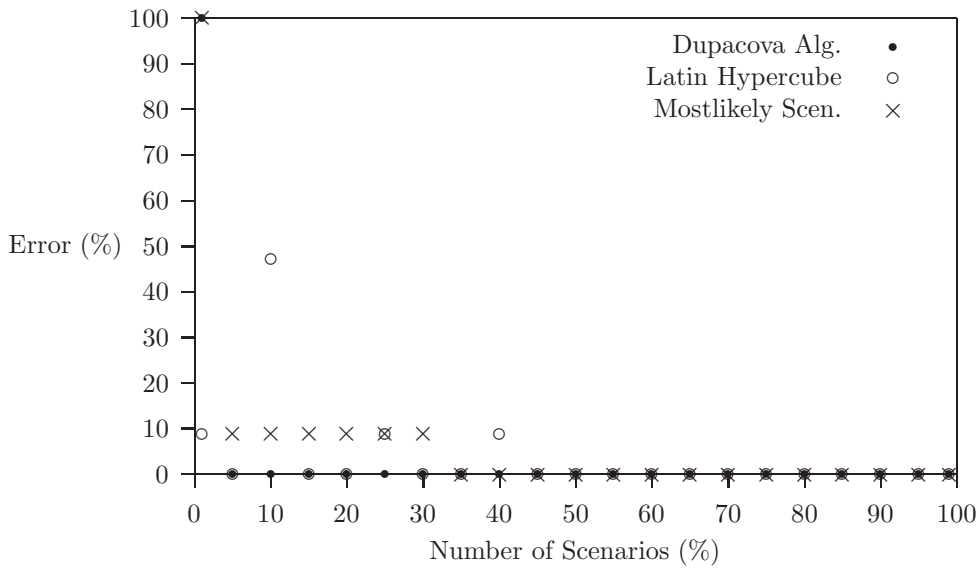

Uncertainty plays a major role in production and inventory
management. In this simplified production/inventory planning
example, there is a single product, a single stocking point,
production capacity constraints and stochastic demand. The objective
is to find the minimum expected cost policy. The cost components
taken into account are holding costs, backlogging costs, fixed
replenishment (or setup) costs and unit production costs. The
optimal policy gives the timing of the replenishments as well as the
order-up-to-levels. Hence, the exact order quantity can be known
only after the realization of the demand, using the scenario
dependent order-up-to-level decisions. This test problem has 5
stages (4 states in each) and 1,024 scenarios. The CP model has 26
decision variables and 21 constraints for one scenario, and 4,775
decision variables and 3,411 constraints for 1,024 scenarios. In Fig.
\ref{fig6}, we again see that Dupacova et al's algorithm and Latin
hypercube sampling are very effective, but both are now only a small
distance ahead of the most likely scenario method.

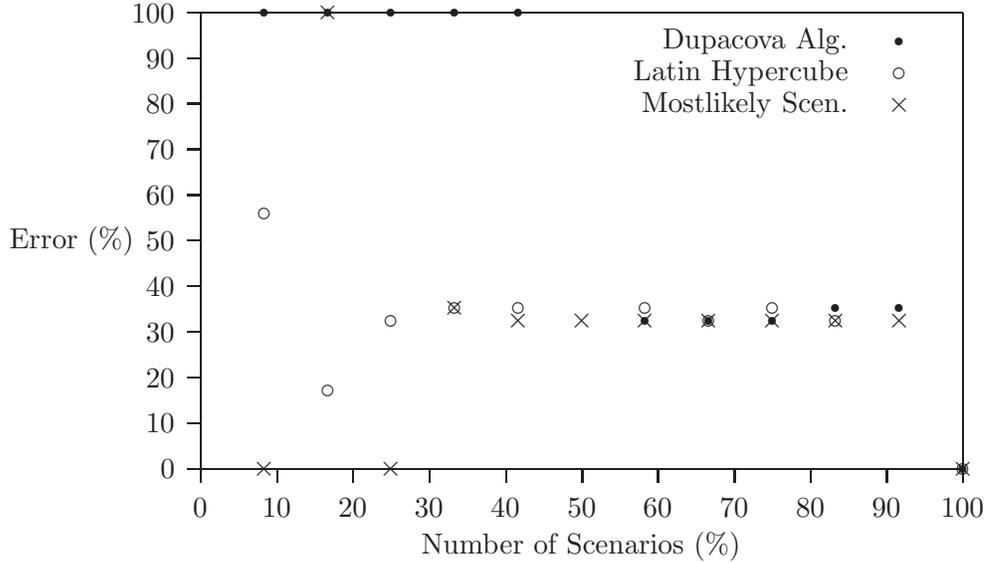
\begin{figure}[h]
\begin{center}
\input{chance.tex}
\end{center}
\caption{Production/Inventory Management (Service Level)}
\label{fig7}
\end{figure}

The study of stochastic production/inventory problems is often
divided into two broad groups: the full cost model and the partial
cost model with a service level constraint. An example of the full
cost model is given above. The service level approach introduces a
service level constraint in place of the shortage cost, where the
service level refers to the availability of stock in an expected
sense. A certainty equivalent MIP formulation of this problem,
under non-stationary stochastic continuous demand assumption, is
given by \cite{tar04}. The same problem, but under discrete demand
assumption, is tackled here using the Stochastic CP framework and
the results for the scenario reduction algorithms are given in
Fig. \ref{fig7}.

The performances of scenario reduction algorithms exhibit a
completely different pattern in the service-level version of the
production/inventory example. In contrast to our previous
observations, now the ``mostlikely'' method outperforms other two
methods. In only 1 case out of a total of 11 cases the
``mostlikely'' method is surpassed by LHS. It should also be noted
that Dupacova et al. method gives an infeasible solution in one
case and LHS in two cases, whereas the ``mostlikely'' method
consistently produces feasible solutions. An explanation for this
outcome lies in the very nature of the problem under consideration
and the probability redistribution mechanism of the scenario
reduction algorithms. In Dupacova et al., in each iteration two
closest scenarios are chosen and reduced to one. The deleted
scenario's probability is added to the preserved one's
probability. Likewise, in LHS, probabilities are redistributed in
accordance with the outcome of simulation experiments. These two
methods modify the existing probability structure substantially,
whereas the ``mostlikely'' method chooses the most probable
scenarios and then only normalises their probabilities, which is
less of a radical change compared to other two methods. Therefore,
the better performance of the ``mostlikely'' method hinges on
these aspects of scenario reduction algorithms in conjunction with
those of chance-constrained problems where probabilities are not
only a factor that affects the expected value of the objective
function but feasibility itself.

\section{Robust solutions \label{robust}}

Inspired by robust optimization methods in operations research
\cite{kouvelis96}, we can also find robust
solutions to stochastic constraint programs. That is, solutions in
which similar decisions are made in different scenarios. It
will often be impossible or undesirable for all decision variables
to be robust. We therefore identify those decision variables whose
values we wish to be identical across scenarios using commands of
the form:
\begin{verbatim}
  robust <Var>;
\end{verbatim}

For example, in production/inventory problem of Sec.\ref{proinv}
the decision variables ``order-up-to-levels'' and ``replenishment
periods'' can be declared as robust variables. The values of these
two sets of decision variables are then fixed at the beginning of
the planning horizon giving a static policy. A robust solution
dampens the nervousness of the solution, an area of very active
research in production/inventory management. As the expected cost
of the robust solution is always higher, the tradeoff between
nervousness and cost may have to be taken into account.

According to \cite{mul95}, the optimal solution of the program
will be robust with respect to optimality if it remains close to
optimal for any realization of the scenario $\omega \in \Omega$.
It is then termed ``solution robust''. The objective function can
be written in the form, $\min \sigma (x,y_{\omega \in \Omega})$
where $x$ denotes the deterministic decision variables, $y_{\omega
\in \Omega}$ is a set of control variables for each scenario.

There is not a unique form of the above function. As discussed in
Sec.\ref{obj_func}, one typical form can be the expected value
criterion, $\sigma(.)=\sum_{\omega \in \Omega} p_\omega Q_\omega$,
in which the objective function of a model becomes a random
variable taking the value $Q_\omega$ with probability $p_\omega$.
Another common form is the worst-case criterion,
$\sigma(.)=\max_{\omega \in \Omega} Q_\omega$.

Mulvey, Vanderbei and Zenios point out that the expected value and
the worst-case functions are special cases in robust optimization,
and the tradeoff between mean value and its variability is a
novelty of the robust optimization formulation. However, as
discussed in Sec.\ref{obj_func}, Markowitz's mean/variance model
provides such a framework.

To demonstrate the concept and the use of
robustness in stochastic constraint programming,
we consider a production/inventory planning problem with demand
data provided in Table \ref{data}, a production capacity of 40
units/period, and stationary costs: production/purchasing costs
\$2/unit, fixed ordering (or setup) costs \$50/replenishment,
inventory holding costs \$1/unit/period, backlogging costs
\$5/unit/period.
\begin{table}[h]
\begin{center}
\begin{tabular}{c|cccc|cccc}
  \hline
   {} & $D_1$ & $D_2$ & $D_3$ & $D_4$ & Pr\{$D_1$\} &Pr\{$D_2$\}&Pr\{$D_3$\}&Pr\{$D_4$\} \\
  \hline
  Period 1 & 8 & 10 & 12 & 14 & 0.20 & 0.20 & 0.40 & 0.20\\
  Period 2 & 15 & 18 & 21 & 24 & 0.30 & 0.40 & 0.20 & 0.10\\
  Period 3 & 15 & 20 & 23 & 26 & 0.10 & 0.20 & 0.60 & 0.10\\
  Period 4 & 10 & 15 & 20 & 22 & 0.50 & 0.30 & 0.10 & 0.10\\
  Period 5 & 12 & 18 & 20 & 24 & 0.30 & 0.50 & 0.10 & 0.10\\
  \hline
\end{tabular}
\caption{Demand data \label{data}}
\end{center}
\end{table}

Stochastic-CP model, with \verb"minimize expected(cost)", gives
the following solution: In Period 1, inventory is raised to 33; no
replenishment is planned for Period 2; in Period 3, a
replenishment is planned with a scenario dependent
order-up-to-level varying between 35 to 47; in Period 4, only in
one scenario out of 64 there is a replenishment; in the final
period, there are replenishments with various order-up-to-levels
in 36 scenarios out of 256. The expected total cost of following
the optimal policy is \$351.61.

To have a static policy, order-up-to-levels and replenishment
decisions are declared as

\verb"robust int+ Order_up_to[Period] in 0..maxint;"

and

\verb"robust int Replenish[Period] in 0..1;"

which give a less nervous policy. Now the best replenishment
policy is to have replenishment every period, Periods 1-5, with an
order-up-to-level [14,21,23,20,18] respectively. The expected
total cost of this scheme is \$439.70.

The solution robust policy, assuming a tradeoff constant $\lambda
= 1.0$, is determined using the Stochastic OPL objective function
\verb"minimize mv(cost,"$\lambda$\verb")". The optimal
replenishment plan is now: in Period 1, the order-up-to-level is
35; no replenishment is planned for Period 2; in Period 3,
depending on demand, order-up-to-level takes a value between 37
and 51; no replenishment in Period 4; and finally, in Period 5,
there are replenishments in 16 scenarios. The expected total cost
of following the solution robust policy is \$353.06.

\begin{figure}[h]
\begin{center}
\input{risk.tex}
\end{center}
\caption{$\lambda$ vs. Risk, Prod/Inv Cost, Obj.Func Value
\label{risk}}
\end{figure}
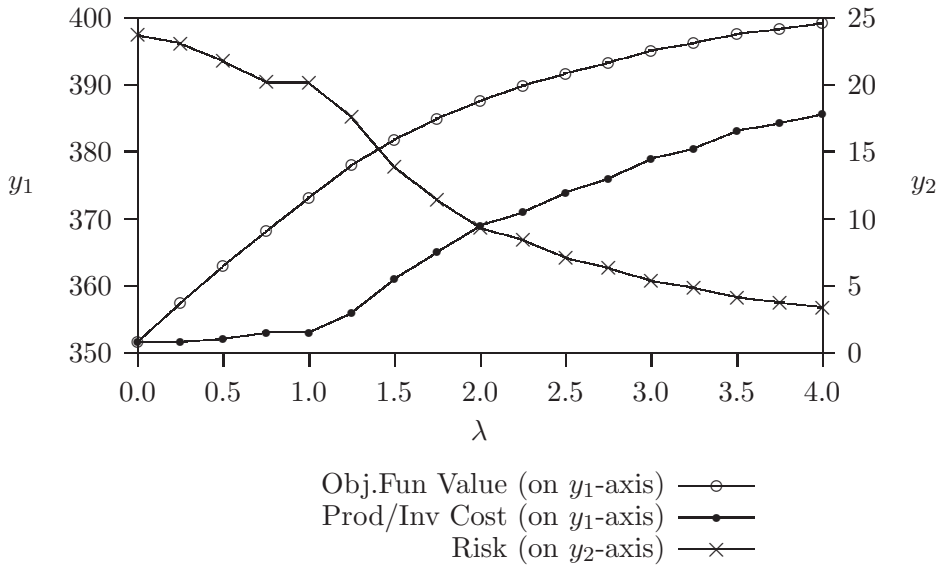

In Fig.\ref{risk}, we see that as one would expect, an increase in
$\lambda$, which actually points to a decrease in the objective
value uncertainty, causes an increase in the expected total
production and inventory costs.

\section{Related work in decision making under uncertainty}

Stochastic constraint programs are closely related to Markov
decision problems (MDPs). An MDP model consists of a set of
states, a set of actions, a state transition function which gives
the probability of moving between two states as a result of a
given action, and a reward function. A solution to an MDP is a
policy, which specifies the best action to take in each possible
state. MDPs have been very influential in AI of late for dealing
with situations involving reasoning under uncertainty
\cite{puterman1}. Stochastic constraint programs can model
problems which lack the Markov property that the next state and
reward depend only on the previous state and action taken. To
represent a stochastic constraint program in which the current
decision depends on all earlier decisions would require an MDP
with an exponential number of states. Stochastic constraint
optimization can also be used to model more complex reward
functions than the (discounted) sum of individual rewards. Another
significant difference is that stochastic constraint programs by
using a scenario-based interpretation can immediately call upon
complex and powerful constraint propagation techniques.

Stochastic constraint programs are also closely related to
influence diagrams. Influence diagrams are Bayesian networks in
which the chance nodes are augmented with decision and utility
nodes \cite{oliver1}. The usual aim is to maximize the sum of the
expected utilities. Chance nodes in an influence diagram
correspond to stochastic variables in a stochastic constraint
program, whilst decision nodes correspond to decision variables.
The utility nodes correspond to the cost function in a stochastic
constraint optimization problem.
However, reasoning about stochastic constraint
programs is likely to be easier than about influence diagrams.
First, the probabilistic aspect of a stochastic constraint program
is simple and decomposable as there are only unary marginal
probabilities. Second, the dependencies between decision variables
and stochastic variables are represented by declarative
constraints. We can therefore borrow from traditional constraint
satisfaction and optimization powerful algorithmic techniques like
branch and bound, constraint propagation and nogood recording. As
a result, if a problem can be modelled within the more restricted
format of a stochastic constraint program, we hope to be able to
reason about it more efficiently.

\section{Related work in constraints}

Stochastic constraint programming was inspired by both stochastic
integer programming and stochastic satisfiability
\cite{littman00}. It is designed to take advantage of some of the
best features of each framework. For example, we are able to write
expressive models using non-linear and global constraints, and to
exploit efficient constraint propagation algorithms. In operations
research, scenarios are used in stochastic programming. Indeed,
the scenario reduction techniques of Dupacova, Growe-Kuska and
Romisch \cite{dupacova03} implemented here are borrowed directly
from stochastic programming.

Mixed constraint satisfaction \cite{fargier2} is closely related
to one stage stochastic constraint satisfaction. In a mixed CSP,
the decision variables are set after the stochastic variables are
given random values. In addition, the random values are chosen
uniformly. In the case of full observability, the aim is to find
conditional values for the decision variables in a mixed CSP so
that we satisfy all possible worlds. In the case of no
observability, the aim is to find values for the decision
variables in a mixed CSP so that we satisfy as many possible
worlds. An earlier constraint satisfaction model for decision
making under uncertainty \cite{fargier3} also included a
probability distribution over the space of possible worlds.

Constraint satisfaction has been extended to include probabilistic
preferences on the values assigned to variables \cite{shazeer1}.
Associated with the values for each variable is a probability
distribution. A ``best'' solution to the constraint satisfaction
problem is then found. This may be the maximum probability
solution (which satisfies the constraints and is most probable),
or the maximum expected overlap solution (which is most like the
true solution).
The latter 
can be viewed as the solution which has the maximum expected
overlap with one generated at random using the probability
distribution. The maximum expected overlap solution could be found
by solving a suitable one stage stochastic constraint optimization
problem.

Branching constraint satisfaction \cite{fowler1} models problems
in which there is uncertainty in the number of variables. For
example, we can model a nurse rostering problem by assigning
shifts to nurses. Branching constraint satisfaction then allows us
to deal with the uncertainty in which nurses are available for
duty. We can represent such problems with a stochastic
CSP with a stochastic 0/1 variable
for each nurse representing their availability. 

A number of extensions of the traditional constraint satisfaction
problem model constraints that are uncertain, probabilistic or not
necessarily satisfied (see, for instance,
\cite{fargier1,schiex5,ysg2003}). In partial constraint
satisfaction we maximize the number of constraints satisfied
\cite{freuder5}. As a second example, in probabilistic constraint
satisfaction each constraint has a certain probability independent
of all other probabilities of being part of the problem
\cite{fargier1}. As a third example, both valued and semi-ring
based constraint satisfaction \cite{schiex5} generalizes
probabilistic constraint satisfaction as well as a number of other
frameworks. In semi-ring based constraint satisfaction, a value is
associated with each tuple in a constraint, whilst in valued
constraint satisfaction, a value is associated with each
constraint. As a fourth example, the certainty closure model
\cite{ysg2003} permits constraints to have parameters whose values
are uncertain. This differs from stochastic constraint programming
in three significant ways. First, stochastic variables come with
probability distributions in our framework, whilst the uncertain
parameters in the certainty closure model take any of their
possible values. Second, we find a policy which may react
differently according to the values taken by the stochastic
variables, whilst the certainty closure model aims to find the
decision space within which all possible solutions must be
contained. Third, stochastic constraint programs can have multiple
stages whilst certainty closure models are essentially one stage.
Stochastic constraint programming can easily be combined with most
of these techniques. For example, we can define stochastic partial
constraint satisfaction in which we maximize the number of
satisfied constraints, or stochastic probabilistic constraint
satisfaction in which each constraint has an associated
probability of being in the problem.

\section{Conclusions}

We have described stochastic constraint programming, an extension
of constraint programming to deal with both decision variables
(which we can set) and stochastic variables (which follow some
probability distribution). This framework is designed to take
advantage of the best features of traditional constraint
satisfaction, stochastic integer programming, and stochastic
satisfiability. It can be used to model a wide variety of decision
problems involving uncertainty and probability. We have
provided a semantics for stochastic constraint
programs based on scenarios.
We have shown how to compile stochastic constraint programs down into
conventional (non-stochastic) constraint programs.
We can therefore call upon the full power of existing
constraint solvers without any modification. We have also
described a number of techniques to reduce the number of
scenarios, and to generate robust solutions.

We have implemented this framework for decision making under
uncertainty in a language called stochastic OPL. This is an
extension of the OPL constraint modelling language \cite{hent99}.
To illustrate the potential of this framework, we have modelled a
wide range of problems in areas as diverse as finance, agriculture
and production. There are many directions for future work. For
example, we want to allow the user to define a limited set of
scenarios that are representative of the whole. As a second
example, we want to explore more sophisticated notions of solution
robustness (e.g. limiting the range of values used by a decision
variable).

\section*{Acknowledgements}

S. Armagan Tarim is supported by Science Foundation Ireland under Grant No. 03/CE3/I405
as part of the Centre for Telecommunications Value-­Chain-­Driven
Research (CTVR) and Grant No. 00/PI.1/C075, and by EPSRC under GR/R30792.
Suresh Manandhar is supported by EPSRC under GR/R30792. We thank
the other members of the 4C Lab and APES Research Group.

\section*{Appendix}

In this appendix, we present a mathematical formulation of the
portfolio diversification problem of Section \ref{portfolio}. We
also give corresponding stochastic and certainty-equivalent OPL
representations.

Assume that, we have $\$W$ to invest, denoted by decision variables $inv[i]$, in any of $i \in I$ instruments (stocks, bonds, etc.)
over a planning horizon of $N$ periods,
\begin{eqnarray}
&&wealth[1] = W \nonumber\\
&&wealth[p] = \sum_{i \in I} inv[i,p], \hspace{1em} p \in \{1,...,N\} \nonumber
\end{eqnarray}
We wish to exceed a wealth of $\$G$ at the end of the planning horizon.
To calculate the utility, we suppose that
exceeding $\$G$ is equivalent to an income of $q\%$ of the excess
while not meeting the goal is equivalent to borrowing at a cost
$r\%$ of the amount short. So, we can write,
\begin{eqnarray}
&&\max E(q\times pos - r \times neg) \nonumber\\
&&pos - neg = wealth[N+1] - G \nonumber\\
&&pos, neg \in \mathbb{Z}^{0,+}. \nonumber
\end{eqnarray}
The uncertainty in this problem is the rate of return, $return$,
which is a random variable, on each investment in each period.
\begin{equation}
wealth[p+1] \leq \sum_{i \in I} inv[i,p](1+return[i,p]), \hspace{1em} p \in \{1,...,N\} \nonumber
\end{equation}
A stochastic CP model for the above formulation, written in stochastic OPL, is as follows:
\begin{verbatim}
stoch market[Period] = ...;
enum I ...;
int N = ...;
range Period [1..N];
range Period_M [1..N+1];
float return[I,Period]^market = ...;
var int+ inv[I,Period] in 0..200;
var int wealth[Period_M] in 100..200;
var int+ pos in 0..50;
var int+ neg in 0..50;

maximize expected(pos - 4*neg)
subject to {
wealth[1] = 100;
pos - neg = wealth[N+1] - 150;
forall (p in Period) wealth[p] = sum (i in I) inv[i,p];
forall (p in Period) wealth[p+1] <= sum (i in I) inv[i,p]*(1+return[i,p]);
};
\end{verbatim}
A possible corresponding input data is
\begin{verbatim}
I = {stock,bond};
N = 3;
market =[<0.0(0.5),0.0(0.5)>,<0.0(0.5),0.0(0.5)>,<0.0(0.5),0.0(0.5)>];
return = [[<0.25,0.06>,<0.25,0.06>,<0.25,0.06>],[<0.14,0.12>,<0.14,0.12>,<0.14,0.12>]];
\end{verbatim}

The stochastic OPL model can then be compiled down into the following certainty-equivalent OPL model:
\begin{verbatim}
enum I ...;
int N = ...;
range Period_Ext [1..N+1];
range Period [1..N];
range Period_M [1..N+1];
int nbStates[Period] = ...;
int nbNodes[Period_Ext] = ...;
int+ ScenTree[Period_Ext,1..nbNodes[N+1]] = ...;
struct TreeType {Period_Ext stage; int state; };
{TreeType} States ={<stage,state>|stage in Period & state in 1..nbStates[stage]};
{TreeType} Nodes ={<stage,state>|stage in Period & state in 1..nbNodes[stage]};
{TreeType} Nodes_Ext ={<stage,state>|stage in Period_Ext & state in 1..nbNodes[stage]};
{TreeType} Nodes_M ={<stage,state>|stage in Period_M & state in 1..nbNodes[stage]};
float Probability[Nodes_Ext] = ...;
float market[States] = ...;
float return[I,States] = ... ;
var int+ inv[I,Nodes] in 0..maxint;
var int wealth[Nodes_M] in 0..maxint;
var int+ pos[1..nbNodes[N+1]] in 0..maxint;
var int+ neg[1..nbNodes[N+1]] in 0..maxint;

maximize sum(scen in 1..nbNodes[N+1]) (pos[scen]-4*neg[scen])*Probability[<N+1,scen>]
subject to {
forall(scen in 1..nbNodes[N+1]) {
wealth[<1,ScenTree[1,scen]>] = 100;
pos[scen]-neg[scen] = wealth[<N+1,ScenTree[N+1,scen]>]-150;
forall(p in Period) wealth[<p,ScenTree[p,scen]>] =
    sum(i in I) inv[i,<p,ScenTree[p,scen]>];
forall(p in Period) wealth[<p+1,ScenTree[p+1,scen]>] <=
    sum(i in I) inv[i,<p,ScenTree[p,scen]>]*(1+return[i,<p,ScenTree[p+1,scen]>]);
};
};
\end{verbatim}
The corresponding problem data is also compiled into
\begin{verbatim}
I={stock,bond};
N=3;
nbStates = [2, 4, 8];
nbNodes = [1, 2, 4, 8];
market = [ 0 0 0 0 0 0 0 0 0 0 0 0 0 0 ];
return = [
[ 0.25 0.06 0.25 0.06 0.25 0.06 0.25 0.06 0.25 0.06 0.25 0.06 0.25 0.06 ]
[ 0.14 0.12 0.14 0.12 0.14 0.12 0.14 0.12 0.14 0.12 0.14 0.12 0.14 0.12 ]];
Probability = [1.0 0.5 0.5 0.25 0.25 0.25 0.25 0.125 0.125 0.125
0.125 0.125 0.125 0.125 0.125 ];
ScenTree = [
[ 1 1 1 1 1 1 1 1 ],
[ 1 1 1 1 2 2 2 2 ],
[ 1 1 2 2 3 3 4 4 ],
[ 1 2 3 4 5 6 7 8 ]];
\end{verbatim}

\begin{figure}[htbp]
\centering \epsfig{file=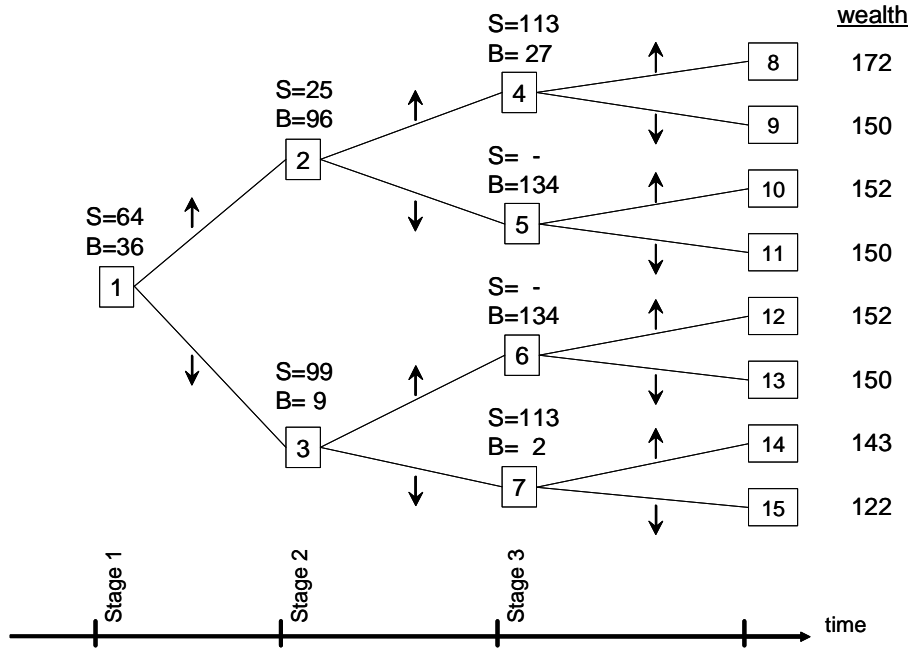, width=12cm} \caption{The
optimal investment plan\label{inv_plan} }
\end{figure}

This problem is solved to optimality and the obtained investment
plan is depicted in Fig.\ref{inv_plan}. $S$ and $B$ denotes the
amounts that should be invested in stocks and bonds, respectively.

\newpage

\bibliographystyle{}

\end{document}

%% file: sb1.tex
\setlength{\unitlength}{0.240900pt}
\ifx\plotpoint\undefined\newsavebox{\plotpoint}\fi
\sbox{\plotpoint}{\rule[-0.200pt]{0.400pt}{0.400pt}}%
\begin{picture}(1500,900)(0,0)
\font\gnuplot=cmr10 at 10pt
\gnuplot
\sbox{\plotpoint}{\rule[-0.200pt]{0.400pt}{0.400pt}}%
\put(202,143){\makebox(0,0)[r]{0}}
\put(222.0,143.0){\rule[-0.200pt]{4.818pt}{0.400pt}}
\put(202,215){\makebox(0,0)[r]{10}}
\put(222.0,215.0){\rule[-0.200pt]{4.818pt}{0.400pt}}
\put(202,286){\makebox(0,0)[r]{20}}
\put(222.0,286.0){\rule[-0.200pt]{4.818pt}{0.400pt}}
\put(202,358){\makebox(0,0)[r]{30}}
\put(222.0,358.0){\rule[-0.200pt]{4.818pt}{0.400pt}}
\put(202,430){\makebox(0,0)[r]{40}}
\put(222.0,430.0){\rule[-0.200pt]{4.818pt}{0.400pt}}
\put(202,501){\makebox(0,0)[r]{50}}
\put(222.0,501.0){\rule[-0.200pt]{4.818pt}{0.400pt}}
\put(202,573){\makebox(0,0)[r]{60}}
\put(222.0,573.0){\rule[-0.200pt]{4.818pt}{0.400pt}}
\put(202,645){\makebox(0,0)[r]{70}}
\put(222.0,645.0){\rule[-0.200pt]{4.818pt}{0.400pt}}
\put(202,717){\makebox(0,0)[r]{80}}
\put(222.0,717.0){\rule[-0.200pt]{4.818pt}{0.400pt}}
\put(202,788){\makebox(0,0)[r]{90}}
\put(222.0,788.0){\rule[-0.200pt]{4.818pt}{0.400pt}}
\put(202,860){\makebox(0,0)[r]{100}}
\put(222.0,860.0){\rule[-0.200pt]{4.818pt}{0.400pt}}
\put(242,82){\makebox(0,0){0}}
\put(242.0,123.0){\rule[-0.200pt]{0.400pt}{4.818pt}}
\put(362,82){\makebox(0,0){10}}
\put(362.0,123.0){\rule[-0.200pt]{0.400pt}{4.818pt}}
\put(481,82){\makebox(0,0){20}}
\put(481.0,123.0){\rule[-0.200pt]{0.400pt}{4.818pt}}
\put(601,82){\makebox(0,0){30}}
\put(601.0,123.0){\rule[-0.200pt]{0.400pt}{4.818pt}}
\put(721,82){\makebox(0,0){40}}
\put(721.0,123.0){\rule[-0.200pt]{0.400pt}{4.818pt}}
\put(841,82){\makebox(0,0){50}}
\put(841.0,123.0){\rule[-0.200pt]{0.400pt}{4.818pt}}
\put(960,82){\makebox(0,0){60}}
\put(960.0,123.0){\rule[-0.200pt]{0.400pt}{4.818pt}}
\put(1080,82){\makebox(0,0){70}}
\put(1080.0,123.0){\rule[-0.200pt]{0.400pt}{4.818pt}}
\put(1200,82){\makebox(0,0){80}}
\put(1200.0,123.0){\rule[-0.200pt]{0.400pt}{4.818pt}}
\put(1319,82){\makebox(0,0){90}}
\put(1319.0,123.0){\rule[-0.200pt]{0.400pt}{4.818pt}}
\put(1439,82){\makebox(0,0){100}}
\put(1439.0,123.0){\rule[-0.200pt]{0.400pt}{4.818pt}}
\put(242.0,143.0){\rule[-0.200pt]{288.357pt}{0.400pt}}
\put(1439.0,143.0){\rule[-0.200pt]{0.400pt}{172.725pt}}
\put(242.0,860.0){\rule[-0.200pt]{288.357pt}{0.400pt}}
\put(40,501){\makebox(0,0){Error (\%)}}
\put(840,21){\makebox(0,0){Number of Scenarios (\%)}}
\put(242.0,143.0){\rule[-0.200pt]{0.400pt}{172.725pt}}
\sbox{\plotpoint}{\rule[-0.500pt]{1.000pt}{1.000pt}}%
\put(1259,815){\makebox(0,0)[r]{Dupacova Alg.}}
\put(1427,143){\circle*{12}}
\put(1379,143){\circle*{12}}
\put(1319,143){\circle*{12}}
\put(1259,143){\circle*{12}}
\put(1200,143){\circle*{12}}
\put(1140,143){\circle*{12}}
\put(1080,143){\circle*{12}}
\put(1020,143){\circle*{12}}
\put(960,144){\circle*{12}}
\put(900,144){\circle*{12}}
\put(841,146){\circle*{12}}
\put(829,146){\circle*{12}}
\put(817,147){\circle*{12}}
\put(805,146){\circle*{12}}
\put(793,147){\circle*{12}}
\put(781,148){\circle*{12}}
\put(769,150){\circle*{12}}
\put(757,150){\circle*{12}}
\put(745,150){\circle*{12}}
\put(733,156){\circle*{12}}
\put(721,157){\circle*{12}}
\put(709,157){\circle*{12}}
\put(697,157){\circle*{12}}
\put(685,157){\circle*{12}}
\put(673,157){\circle*{12}}
\put(661,162){\circle*{12}}
\put(649,162){\circle*{12}}
\put(637,162){\circle*{12}}
\put(625,164){\circle*{12}}
\put(613,164){\circle*{12}}
\put(601,164){\circle*{12}}
\put(589,164){\circle*{12}}
\put(577,164){\circle*{12}}
\put(565,163){\circle*{12}}
\put(553,163){\circle*{12}}
\put(541,166){\circle*{12}}
\put(529,164){\circle*{12}}
\put(517,164){\circle*{12}}
\put(505,167){\circle*{12}}
\put(493,167){\circle*{12}}
\put(481,174){\circle*{12}}
\put(469,168){\circle*{12}}
\put(457,167){\circle*{12}}
\put(445,164){\circle*{12}}
\put(434,164){\circle*{12}}
\put(422,163){\circle*{12}}
\put(410,163){\circle*{12}}
\put(398,163){\circle*{12}}
\put(386,163){\circle*{12}}
\put(374,163){\circle*{12}}
\put(362,162){\circle*{12}}
\put(350,155){\circle*{12}}
\put(338,162){\circle*{12}}
\put(326,149){\circle*{12}}
\put(314,149){\circle*{12}}
\put(302,143){\circle*{12}}
\put(290,151){\circle*{12}}
\put(278,155){\circle*{12}}
\put(266,232){\circle*{12}}
\put(254,158){\circle*{12}}
\put(1339,815){\circle*{12}}
\sbox{\plotpoint}{\rule[-0.200pt]{0.400pt}{0.400pt}}%
\put(1259,765){\makebox(0,0)[r]{Latin Hypercube}}
\put(1427,143){\circle{18}}
\put(1379,162){\circle{18}}
\put(1319,170){\circle{18}}
\put(1259,144){\circle{18}}
\put(1200,145){\circle{18}}
\put(1140,157){\circle{18}}
\put(1080,159){\circle{18}}
\put(1020,163){\circle{18}}
\put(960,160){\circle{18}}
\put(900,171){\circle{18}}
\put(841,156){\circle{18}}
\put(829,157){\circle{18}}
\put(817,194){\circle{18}}
\put(805,180){\circle{18}}
\put(793,161){\circle{18}}
\put(781,148){\circle{18}}
\put(769,167){\circle{18}}
\put(757,155){\circle{18}}
\put(745,171){\circle{18}}
\put(733,143){\circle{18}}
\put(721,156){\circle{18}}
\put(709,157){\circle{18}}
\put(697,218){\circle{18}}
\put(685,158){\circle{18}}
\put(673,143){\circle{18}}
\put(661,158){\circle{18}}
\put(649,146){\circle{18}}
\put(637,219){\circle{18}}
\put(625,151){\circle{18}}
\put(613,158){\circle{18}}
\put(601,155){\circle{18}}
\put(589,156){\circle{18}}
\put(577,186){\circle{18}}
\put(565,146){\circle{18}}
\put(553,230){\circle{18}}
\put(541,157){\circle{18}}
\put(529,229){\circle{18}}
\put(517,162){\circle{18}}
\put(505,156){\circle{18}}
\put(493,170){\circle{18}}
\put(481,151){\circle{18}}
\put(469,195){\circle{18}}
\put(457,171){\circle{18}}
\put(445,306){\circle{18}}
\put(434,160){\circle{18}}
\put(422,148){\circle{18}}
\put(410,147){\circle{18}}
\put(398,205){\circle{18}}
\put(386,238){\circle{18}}
\put(374,151){\circle{18}}
\put(362,170){\circle{18}}
\put(350,163){\circle{18}}
\put(338,146){\circle{18}}
\put(326,163){\circle{18}}
\put(314,371){\circle{18}}
\put(302,228){\circle{18}}
\put(290,158){\circle{18}}
\put(278,147){\circle{18}}
\put(266,167){\circle{18}}
\put(254,860){\circle{18}}
\put(1339,765){\circle{18}}
\sbox{\plotpoint}{\rule[-0.500pt]{1.000pt}{1.000pt}}%
\put(1259,715){\makebox(0,0)[r]{Mostlikely Scen.}}
\put(1427,143){\makebox(0,0){$\times$}}
\put(1379,143){\makebox(0,0){$\times$}}
\put(1319,143){\makebox(0,0){$\times$}}
\put(1259,143){\makebox(0,0){$\times$}}
\put(1200,143){\makebox(0,0){$\times$}}
\put(1140,144){\makebox(0,0){$\times$}}
\put(1080,146){\makebox(0,0){$\times$}}
\put(1020,150){\makebox(0,0){$\times$}}
\put(960,159){\makebox(0,0){$\times$}}
\put(900,167){\makebox(0,0){$\times$}}
\put(841,186){\makebox(0,0){$\times$}}
\put(829,209){\makebox(0,0){$\times$}}
\put(817,224){\makebox(0,0){$\times$}}
\put(805,229){\makebox(0,0){$\times$}}
\put(793,229){\makebox(0,0){$\times$}}
\put(781,230){\makebox(0,0){$\times$}}
\put(769,236){\makebox(0,0){$\times$}}
\put(757,240){\makebox(0,0){$\times$}}
\put(745,256){\makebox(0,0){$\times$}}
\put(733,289){\makebox(0,0){$\times$}}
\put(721,330){\makebox(0,0){$\times$}}
\put(709,356){\makebox(0,0){$\times$}}
\put(697,370){\makebox(0,0){$\times$}}
\put(685,380){\makebox(0,0){$\times$}}
\put(673,442){\makebox(0,0){$\times$}}
\put(661,495){\makebox(0,0){$\times$}}
\put(649,523){\makebox(0,0){$\times$}}
\put(637,547){\makebox(0,0){$\times$}}
\put(625,577){\makebox(0,0){$\times$}}
\put(613,623){\makebox(0,0){$\times$}}
\put(601,741){\makebox(0,0){$\times$}}
\put(589,826){\makebox(0,0){$\times$}}
\put(577,860){\makebox(0,0){$\times$}}
\put(565,860){\makebox(0,0){$\times$}}
\put(553,860){\makebox(0,0){$\times$}}
\put(541,860){\makebox(0,0){$\times$}}
\put(529,860){\makebox(0,0){$\times$}}
\put(517,860){\makebox(0,0){$\times$}}
\put(505,860){\makebox(0,0){$\times$}}
\put(493,860){\makebox(0,0){$\times$}}
\put(481,860){\makebox(0,0){$\times$}}
\put(469,860){\makebox(0,0){$\times$}}
\put(457,860){\makebox(0,0){$\times$}}
\put(445,860){\makebox(0,0){$\times$}}
\put(434,860){\makebox(0,0){$\times$}}
\put(422,860){\makebox(0,0){$\times$}}
\put(410,860){\makebox(0,0){$\times$}}
\put(398,860){\makebox(0,0){$\times$}}
\put(386,860){\makebox(0,0){$\times$}}
\put(374,860){\makebox(0,0){$\times$}}
\put(362,860){\makebox(0,0){$\times$}}
\put(350,860){\makebox(0,0){$\times$}}
\put(338,860){\makebox(0,0){$\times$}}
\put(326,860){\makebox(0,0){$\times$}}
\put(314,860){\makebox(0,0){$\times$}}
\put(302,860){\makebox(0,0){$\times$}}
\put(290,860){\makebox(0,0){$\times$}}
\put(278,860){\makebox(0,0){$\times$}}
\put(266,860){\makebox(0,0){$\times$}}
\put(254,860){\makebox(0,0){$\times$}}
\put(1339,715){\makebox(0,0){$\times$}}
\end{picture}

%% file: sb2.tex
\setlength{\unitlength}{0.240900pt}
\ifx\plotpoint\undefined\newsavebox{\plotpoint}\fi
\sbox{\plotpoint}{\rule[-0.200pt]{0.400pt}{0.400pt}}%
\begin{picture}(1500,900)(0,0)
\font\gnuplot=cmr10 at 10pt
\gnuplot
\sbox{\plotpoint}{\rule[-0.200pt]{0.400pt}{0.400pt}}%
\put(202,143){\makebox(0,0)[r]{0}}
\put(222.0,143.0){\rule[-0.200pt]{4.818pt}{0.400pt}}
\put(202,215){\makebox(0,0)[r]{10}}
\put(222.0,215.0){\rule[-0.200pt]{4.818pt}{0.400pt}}
\put(202,286){\makebox(0,0)[r]{20}}
\put(222.0,286.0){\rule[-0.200pt]{4.818pt}{0.400pt}}
\put(202,358){\makebox(0,0)[r]{30}}
\put(222.0,358.0){\rule[-0.200pt]{4.818pt}{0.400pt}}
\put(202,430){\makebox(0,0)[r]{40}}
\put(222.0,430.0){\rule[-0.200pt]{4.818pt}{0.400pt}}
\put(202,501){\makebox(0,0)[r]{50}}
\put(222.0,501.0){\rule[-0.200pt]{4.818pt}{0.400pt}}
\put(202,573){\makebox(0,0)[r]{60}}
\put(222.0,573.0){\rule[-0.200pt]{4.818pt}{0.400pt}}
\put(202,645){\makebox(0,0)[r]{70}}
\put(222.0,645.0){\rule[-0.200pt]{4.818pt}{0.400pt}}
\put(202,717){\makebox(0,0)[r]{80}}
\put(222.0,717.0){\rule[-0.200pt]{4.818pt}{0.400pt}}
\put(202,788){\makebox(0,0)[r]{90}}
\put(222.0,788.0){\rule[-0.200pt]{4.818pt}{0.400pt}}
\put(202,860){\makebox(0,0)[r]{100}}
\put(222.0,860.0){\rule[-0.200pt]{4.818pt}{0.400pt}}
\put(242,82){\makebox(0,0){0}}
\put(242.0,123.0){\rule[-0.200pt]{0.400pt}{4.818pt}}
\put(362,82){\makebox(0,0){10}}
\put(362.0,123.0){\rule[-0.200pt]{0.400pt}{4.818pt}}
\put(481,82){\makebox(0,0){20}}
\put(481.0,123.0){\rule[-0.200pt]{0.400pt}{4.818pt}}
\put(601,82){\makebox(0,0){30}}
\put(601.0,123.0){\rule[-0.200pt]{0.400pt}{4.818pt}}
\put(721,82){\makebox(0,0){40}}
\put(721.0,123.0){\rule[-0.200pt]{0.400pt}{4.818pt}}
\put(841,82){\makebox(0,0){50}}
\put(841.0,123.0){\rule[-0.200pt]{0.400pt}{4.818pt}}
\put(960,82){\makebox(0,0){60}}
\put(960.0,123.0){\rule[-0.200pt]{0.400pt}{4.818pt}}
\put(1080,82){\makebox(0,0){70}}
\put(1080.0,123.0){\rule[-0.200pt]{0.400pt}{4.818pt}}
\put(1200,82){\makebox(0,0){80}}
\put(1200.0,123.0){\rule[-0.200pt]{0.400pt}{4.818pt}}
\put(1319,82){\makebox(0,0){90}}
\put(1319.0,123.0){\rule[-0.200pt]{0.400pt}{4.818pt}}
\put(1439,82){\makebox(0,0){100}}
\put(1439.0,123.0){\rule[-0.200pt]{0.400pt}{4.818pt}}
\put(242.0,143.0){\rule[-0.200pt]{288.357pt}{0.400pt}}
\put(1439.0,143.0){\rule[-0.200pt]{0.400pt}{172.725pt}}
\put(242.0,860.0){\rule[-0.200pt]{288.357pt}{0.400pt}}
\put(40,501){\makebox(0,0){Error (\%)}}
\put(840,21){\makebox(0,0){Number of Scenarios (\%)}}
\put(242.0,143.0){\rule[-0.200pt]{0.400pt}{172.725pt}}
\sbox{\plotpoint}{\rule[-0.500pt]{1.000pt}{1.000pt}}%
\put(1259,815){\makebox(0,0)[r]{Dupacova Alg.}}
\put(1427,143){\circle*{12}}
\put(1379,143){\circle*{12}}
\put(1319,143){\circle*{12}}
\put(1259,143){\circle*{12}}
\put(1200,143){\circle*{12}}
\put(1140,143){\circle*{12}}
\put(1080,143){\circle*{12}}
\put(1020,143){\circle*{12}}
\put(960,143){\circle*{12}}
\put(900,143){\circle*{12}}
\put(841,143){\circle*{12}}
\put(829,143){\circle*{12}}
\put(817,143){\circle*{12}}
\put(805,143){\circle*{12}}
\put(793,143){\circle*{12}}
\put(781,143){\circle*{12}}
\put(769,143){\circle*{12}}
\put(757,143){\circle*{12}}
\put(745,143){\circle*{12}}
\put(733,143){\circle*{12}}
\put(721,143){\circle*{12}}
\put(709,194){\circle*{12}}
\put(697,194){\circle*{12}}
\put(685,194){\circle*{12}}
\put(673,194){\circle*{12}}
\put(661,194){\circle*{12}}
\put(649,194){\circle*{12}}
\put(637,194){\circle*{12}}
\put(625,194){\circle*{12}}
\put(613,194){\circle*{12}}
\put(601,194){\circle*{12}}
\put(589,194){\circle*{12}}
\put(577,296){\circle*{12}}
\put(565,296){\circle*{12}}
\put(553,194){\circle*{12}}
\put(541,194){\circle*{12}}
\put(529,296){\circle*{12}}
\put(517,194){\circle*{12}}
\put(505,296){\circle*{12}}
\put(493,194){\circle*{12}}
\put(481,194){\circle*{12}}
\put(469,296){\circle*{12}}
\put(457,194){\circle*{12}}
\put(445,296){\circle*{12}}
\put(434,153){\circle*{12}}
\put(422,143){\circle*{12}}
\put(410,143){\circle*{12}}
\put(398,153){\circle*{12}}
\put(386,143){\circle*{12}}
\put(374,143){\circle*{12}}
\put(362,143){\circle*{12}}
\put(350,143){\circle*{12}}
\put(338,143){\circle*{12}}
\put(326,143){\circle*{12}}
\put(314,143){\circle*{12}}
\put(302,143){\circle*{12}}
\put(290,422){\circle*{12}}
\put(278,422){\circle*{12}}
\put(266,422){\circle*{12}}
\put(254,422){\circle*{12}}
\put(1339,815){\circle*{12}}
\sbox{\plotpoint}{\rule[-0.200pt]{0.400pt}{0.400pt}}%
\put(1259,765){\makebox(0,0)[r]{Latin Hypercube}}
\put(1427,143){\circle{18}}
\put(1379,143){\circle{18}}
\put(1319,296){\circle{18}}
\put(1259,143){\circle{18}}
\put(1200,193){\circle{18}}
\put(1140,143){\circle{18}}
\put(1080,143){\circle{18}}
\put(1020,143){\circle{18}}
\put(960,170){\circle{18}}
\put(900,296){\circle{18}}
\put(841,296){\circle{18}}
\put(829,296){\circle{18}}
\put(817,194){\circle{18}}
\put(805,296){\circle{18}}
\put(793,265){\circle{18}}
\put(781,143){\circle{18}}
\put(769,408){\circle{18}}
\put(757,143){\circle{18}}
\put(745,167){\circle{18}}
\put(733,422){\circle{18}}
\put(721,296){\circle{18}}
\put(709,143){\circle{18}}
\put(697,321){\circle{18}}
\put(685,321){\circle{18}}
\put(673,143){\circle{18}}
\put(661,143){\circle{18}}
\put(649,194){\circle{18}}
\put(637,422){\circle{18}}
\put(625,296){\circle{18}}
\put(613,265){\circle{18}}
\put(601,143){\circle{18}}
\put(589,248){\circle{18}}
\put(577,167){\circle{18}}
\put(565,143){\circle{18}}
\put(553,842){\circle{18}}
\put(541,143){\circle{18}}
\put(529,321){\circle{18}}
\put(517,143){\circle{18}}
\put(505,153){\circle{18}}
\put(493,167){\circle{18}}
\put(481,282){\circle{18}}
\put(469,321){\circle{18}}
\put(457,833){\circle{18}}
\put(445,143){\circle{18}}
\put(434,422){\circle{18}}
\put(422,422){\circle{18}}
\put(410,248){\circle{18}}
\put(398,417){\circle{18}}
\put(386,422){\circle{18}}
\put(374,296){\circle{18}}
\put(362,422){\circle{18}}
\put(350,143){\circle{18}}
\put(338,422){\circle{18}}
\put(326,143){\circle{18}}
\put(314,422){\circle{18}}
\put(302,377){\circle{18}}
\put(290,417){\circle{18}}
\put(278,860){\circle{18}}
\put(266,422){\circle{18}}
\put(254,422){\circle{18}}
\put(1339,765){\circle{18}}
\sbox{\plotpoint}{\rule[-0.500pt]{1.000pt}{1.000pt}}%
\put(1259,715){\makebox(0,0)[r]{Mostlikely Scen.}}
\put(1427,143){\makebox(0,0){$\times$}}
\put(1379,143){\makebox(0,0){$\times$}}
\put(1319,143){\makebox(0,0){$\times$}}
\put(1259,143){\makebox(0,0){$\times$}}
\put(1200,200){\makebox(0,0){$\times$}}
\put(1140,296){\makebox(0,0){$\times$}}
\put(1080,296){\makebox(0,0){$\times$}}
\put(1020,296){\makebox(0,0){$\times$}}
\put(960,422){\makebox(0,0){$\times$}}
\put(900,422){\makebox(0,0){$\times$}}
\put(841,422){\makebox(0,0){$\times$}}
\put(829,422){\makebox(0,0){$\times$}}
\put(817,422){\makebox(0,0){$\times$}}
\put(805,422){\makebox(0,0){$\times$}}
\put(793,422){\makebox(0,0){$\times$}}
\put(781,422){\makebox(0,0){$\times$}}
\put(769,422){\makebox(0,0){$\times$}}
\put(757,422){\makebox(0,0){$\times$}}
\put(745,422){\makebox(0,0){$\times$}}
\put(733,422){\makebox(0,0){$\times$}}
\put(721,422){\makebox(0,0){$\times$}}
\put(709,422){\makebox(0,0){$\times$}}
\put(697,422){\makebox(0,0){$\times$}}
\put(685,422){\makebox(0,0){$\times$}}
\put(673,422){\makebox(0,0){$\times$}}
\put(661,422){\makebox(0,0){$\times$}}
\put(649,422){\makebox(0,0){$\times$}}
\put(637,422){\makebox(0,0){$\times$}}
\put(625,422){\makebox(0,0){$\times$}}
\put(613,422){\makebox(0,0){$\times$}}
\put(601,422){\makebox(0,0){$\times$}}
\put(589,422){\makebox(0,0){$\times$}}
\put(577,422){\makebox(0,0){$\times$}}
\put(565,422){\makebox(0,0){$\times$}}
\put(553,422){\makebox(0,0){$\times$}}
\put(541,422){\makebox(0,0){$\times$}}
\put(529,422){\makebox(0,0){$\times$}}
\put(517,422){\makebox(0,0){$\times$}}
\put(505,422){\makebox(0,0){$\times$}}
\put(493,422){\makebox(0,0){$\times$}}
\put(481,422){\makebox(0,0){$\times$}}
\put(469,422){\makebox(0,0){$\times$}}
\put(457,422){\makebox(0,0){$\times$}}
\put(445,422){\makebox(0,0){$\times$}}
\put(434,422){\makebox(0,0){$\times$}}
\put(422,422){\makebox(0,0){$\times$}}
\put(410,422){\makebox(0,0){$\times$}}
\put(398,422){\makebox(0,0){$\times$}}
\put(386,422){\makebox(0,0){$\times$}}
\put(374,422){\makebox(0,0){$\times$}}
\put(362,422){\makebox(0,0){$\times$}}
\put(350,422){\makebox(0,0){$\times$}}
\put(338,422){\makebox(0,0){$\times$}}
\put(326,422){\makebox(0,0){$\times$}}
\put(314,422){\makebox(0,0){$\times$}}
\put(302,422){\makebox(0,0){$\times$}}
\put(290,422){\makebox(0,0){$\times$}}
\put(278,422){\makebox(0,0){$\times$}}
\put(266,422){\makebox(0,0){$\times$}}
\put(254,422){\makebox(0,0){$\times$}}
\put(1339,715){\makebox(0,0){$\times$}}
\end{picture}

%% file: farm1.tex
\setlength{\unitlength}{0.240900pt}
\ifx\plotpoint\undefined\newsavebox{\plotpoint}\fi
\sbox{\plotpoint}{\rule[-0.200pt]{0.400pt}{0.400pt}}%
\begin{picture}(1500,900)(0,0)
\font\gnuplot=cmr10 at 10pt
\gnuplot
\sbox{\plotpoint}{\rule[-0.200pt]{0.400pt}{0.400pt}}%
\put(202,143){\makebox(0,0)[r]{0}}
\put(222.0,143.0){\rule[-0.200pt]{4.818pt}{0.400pt}}
\put(202,215){\makebox(0,0)[r]{10}}
\put(222.0,215.0){\rule[-0.200pt]{4.818pt}{0.400pt}}
\put(202,286){\makebox(0,0)[r]{20}}
\put(222.0,286.0){\rule[-0.200pt]{4.818pt}{0.400pt}}
\put(202,358){\makebox(0,0)[r]{30}}
\put(222.0,358.0){\rule[-0.200pt]{4.818pt}{0.400pt}}
\put(202,430){\makebox(0,0)[r]{40}}
\put(222.0,430.0){\rule[-0.200pt]{4.818pt}{0.400pt}}
\put(202,501){\makebox(0,0)[r]{50}}
\put(222.0,501.0){\rule[-0.200pt]{4.818pt}{0.400pt}}
\put(202,573){\makebox(0,0)[r]{60}}
\put(222.0,573.0){\rule[-0.200pt]{4.818pt}{0.400pt}}
\put(202,645){\makebox(0,0)[r]{70}}
\put(222.0,645.0){\rule[-0.200pt]{4.818pt}{0.400pt}}
\put(202,717){\makebox(0,0)[r]{80}}
\put(222.0,717.0){\rule[-0.200pt]{4.818pt}{0.400pt}}
\put(202,788){\makebox(0,0)[r]{90}}
\put(222.0,788.0){\rule[-0.200pt]{4.818pt}{0.400pt}}
\put(202,860){\makebox(0,0)[r]{100}}
\put(222.0,860.0){\rule[-0.200pt]{4.818pt}{0.400pt}}
\put(242,82){\makebox(0,0){0}}
\put(242.0,123.0){\rule[-0.200pt]{0.400pt}{4.818pt}}
\put(362,82){\makebox(0,0){10}}
\put(362.0,123.0){\rule[-0.200pt]{0.400pt}{4.818pt}}
\put(481,82){\makebox(0,0){20}}
\put(481.0,123.0){\rule[-0.200pt]{0.400pt}{4.818pt}}
\put(601,82){\makebox(0,0){30}}
\put(601.0,123.0){\rule[-0.200pt]{0.400pt}{4.818pt}}
\put(721,82){\makebox(0,0){40}}
\put(721.0,123.0){\rule[-0.200pt]{0.400pt}{4.818pt}}
\put(841,82){\makebox(0,0){50}}
\put(841.0,123.0){\rule[-0.200pt]{0.400pt}{4.818pt}}
\put(960,82){\makebox(0,0){60}}
\put(960.0,123.0){\rule[-0.200pt]{0.400pt}{4.818pt}}
\put(1080,82){\makebox(0,0){70}}
\put(1080.0,123.0){\rule[-0.200pt]{0.400pt}{4.818pt}}
\put(1200,82){\makebox(0,0){80}}
\put(1200.0,123.0){\rule[-0.200pt]{0.400pt}{4.818pt}}
\put(1319,82){\makebox(0,0){90}}
\put(1319.0,123.0){\rule[-0.200pt]{0.400pt}{4.818pt}}
\put(1439,82){\makebox(0,0){100}}
\put(1439.0,123.0){\rule[-0.200pt]{0.400pt}{4.818pt}}
\put(242.0,143.0){\rule[-0.200pt]{288.357pt}{0.400pt}}
\put(1439.0,143.0){\rule[-0.200pt]{0.400pt}{172.725pt}}
\put(242.0,860.0){\rule[-0.200pt]{288.357pt}{0.400pt}}
\put(40,501){\makebox(0,0){Error (\%)}}
\put(840,21){\makebox(0,0){Number of Scenarios (\%)}}
\put(242.0,143.0){\rule[-0.200pt]{0.400pt}{172.725pt}}
\sbox{\plotpoint}{\rule[-0.500pt]{1.000pt}{1.000pt}}%
\put(1259,815){\makebox(0,0)[r]{Dupacova Alg.}}
\put(1427,143){\circle*{12}}
\put(1379,143){\circle*{12}}
\put(1319,143){\circle*{12}}
\put(1259,143){\circle*{12}}
\put(1200,143){\circle*{12}}
\put(1140,143){\circle*{12}}
\put(1080,143){\circle*{12}}
\put(1020,143){\circle*{12}}
\put(960,143){\circle*{12}}
\put(900,143){\circle*{12}}
\put(841,143){\circle*{12}}
\put(781,143){\circle*{12}}
\put(721,143){\circle*{12}}
\put(661,143){\circle*{12}}
\put(601,143){\circle*{12}}
\put(541,143){\circle*{12}}
\put(481,143){\circle*{12}}
\put(422,143){\circle*{12}}
\put(362,143){\circle*{12}}
\put(302,143){\circle*{12}}
\put(254,152){\circle*{12}}
\put(1339,815){\circle*{12}}
\sbox{\plotpoint}{\rule[-0.200pt]{0.400pt}{0.400pt}}%
\put(1259,765){\makebox(0,0)[r]{Latin Hypercube}}
\put(1427,143){\circle{18}}
\put(1379,143){\circle{18}}
\put(1319,143){\circle{18}}
\put(1259,143){\circle{18}}
\put(1200,143){\circle{18}}
\put(1140,143){\circle{18}}
\put(1080,143){\circle{18}}
\put(1020,143){\circle{18}}
\put(960,143){\circle{18}}
\put(900,143){\circle{18}}
\put(841,143){\circle{18}}
\put(781,143){\circle{18}}
\put(721,143){\circle{18}}
\put(661,143){\circle{18}}
\put(601,143){\circle{18}}
\put(541,143){\circle{18}}
\put(481,143){\circle{18}}
\put(422,143){\circle{18}}
\put(362,143){\circle{18}}
\put(302,143){\circle{18}}
\put(254,143){\circle{18}}
\put(1427,143){\circle{18}}
\put(1339,765){\circle{18}}
\sbox{\plotpoint}{\rule[-0.500pt]{1.000pt}{1.000pt}}%
\put(1259,715){\makebox(0,0)[r]{Mostlikely Scen.}}
\put(1427,143){\makebox(0,0){$\times$}}
\put(1379,143){\makebox(0,0){$\times$}}
\put(1319,143){\makebox(0,0){$\times$}}
\put(1259,143){\makebox(0,0){$\times$}}
\put(1200,143){\makebox(0,0){$\times$}}
\put(1140,143){\makebox(0,0){$\times$}}
\put(1080,143){\makebox(0,0){$\times$}}
\put(1020,143){\makebox(0,0){$\times$}}
\put(960,143){\makebox(0,0){$\times$}}
\put(900,143){\makebox(0,0){$\times$}}
\put(841,143){\makebox(0,0){$\times$}}
\put(781,143){\makebox(0,0){$\times$}}
\put(721,155){\makebox(0,0){$\times$}}
\put(661,160){\makebox(0,0){$\times$}}
\put(601,287){\makebox(0,0){$\times$}}
\put(541,290){\makebox(0,0){$\times$}}
\put(481,290){\makebox(0,0){$\times$}}
\put(422,290){\makebox(0,0){$\times$}}
\put(362,290){\makebox(0,0){$\times$}}
\put(302,805){\makebox(0,0){$\times$}}
\put(254,860){\makebox(0,0){$\times$}}
\put(1339,715){\makebox(0,0){$\times$}}
\end{picture}

%% file: farm3.tex
\setlength{\unitlength}{0.240900pt}
\ifx\plotpoint\undefined\newsavebox{\plotpoint}\fi
\sbox{\plotpoint}{\rule[-0.200pt]{0.400pt}{0.400pt}}%
\begin{picture}(1500,900)(0,0)
\font\gnuplot=cmr10 at 10pt
\gnuplot
\sbox{\plotpoint}{\rule[-0.200pt]{0.400pt}{0.400pt}}%
\put(202,143){\makebox(0,0)[r]{0}}
\put(222.0,143.0){\rule[-0.200pt]{4.818pt}{0.400pt}}
\put(202,215){\makebox(0,0)[r]{10}}
\put(222.0,215.0){\rule[-0.200pt]{4.818pt}{0.400pt}}
\put(202,286){\makebox(0,0)[r]{20}}
\put(222.0,286.0){\rule[-0.200pt]{4.818pt}{0.400pt}}
\put(202,358){\makebox(0,0)[r]{30}}
\put(222.0,358.0){\rule[-0.200pt]{4.818pt}{0.400pt}}
\put(202,430){\makebox(0,0)[r]{40}}
\put(222.0,430.0){\rule[-0.200pt]{4.818pt}{0.400pt}}
\put(202,501){\makebox(0,0)[r]{50}}
\put(222.0,501.0){\rule[-0.200pt]{4.818pt}{0.400pt}}
\put(202,573){\makebox(0,0)[r]{60}}
\put(222.0,573.0){\rule[-0.200pt]{4.818pt}{0.400pt}}
\put(202,645){\makebox(0,0)[r]{70}}
\put(222.0,645.0){\rule[-0.200pt]{4.818pt}{0.400pt}}
\put(202,717){\makebox(0,0)[r]{80}}
\put(222.0,717.0){\rule[-0.200pt]{4.818pt}{0.400pt}}
\put(202,788){\makebox(0,0)[r]{90}}
\put(222.0,788.0){\rule[-0.200pt]{4.818pt}{0.400pt}}
\put(202,860){\makebox(0,0)[r]{100}}
\put(222.0,860.0){\rule[-0.200pt]{4.818pt}{0.400pt}}
\put(242,82){\makebox(0,0){0}}
\put(242.0,123.0){\rule[-0.200pt]{0.400pt}{4.818pt}}
\put(362,82){\makebox(0,0){10}}
\put(362.0,123.0){\rule[-0.200pt]{0.400pt}{4.818pt}}
\put(481,82){\makebox(0,0){20}}
\put(481.0,123.0){\rule[-0.200pt]{0.400pt}{4.818pt}}
\put(601,82){\makebox(0,0){30}}
\put(601.0,123.0){\rule[-0.200pt]{0.400pt}{4.818pt}}
\put(721,82){\makebox(0,0){40}}
\put(721.0,123.0){\rule[-0.200pt]{0.400pt}{4.818pt}}
\put(841,82){\makebox(0,0){50}}
\put(841.0,123.0){\rule[-0.200pt]{0.400pt}{4.818pt}}
\put(960,82){\makebox(0,0){60}}
\put(960.0,123.0){\rule[-0.200pt]{0.400pt}{4.818pt}}
\put(1080,82){\makebox(0,0){70}}
\put(1080.0,123.0){\rule[-0.200pt]{0.400pt}{4.818pt}}
\put(1200,82){\makebox(0,0){80}}
\put(1200.0,123.0){\rule[-0.200pt]{0.400pt}{4.818pt}}
\put(1319,82){\makebox(0,0){90}}
\put(1319.0,123.0){\rule[-0.200pt]{0.400pt}{4.818pt}}
\put(1439,82){\makebox(0,0){100}}
\put(1439.0,123.0){\rule[-0.200pt]{0.400pt}{4.818pt}}
\put(242.0,143.0){\rule[-0.200pt]{288.357pt}{0.400pt}}
\put(1439.0,143.0){\rule[-0.200pt]{0.400pt}{172.725pt}}
\put(242.0,860.0){\rule[-0.200pt]{288.357pt}{0.400pt}}
\put(40,501){\makebox(0,0){Error (\%)}}
\put(840,21){\makebox(0,0){Number of Scenarios (\%)}}
\put(242.0,143.0){\rule[-0.200pt]{0.400pt}{172.725pt}}
\sbox{\plotpoint}{\rule[-0.500pt]{1.000pt}{1.000pt}}%
\put(1259,815){\makebox(0,0)[r]{Dupacova Alg.}}
\put(1439,143){\circle*{12}}
\put(1319,143){\circle*{12}}
\put(1200,143){\circle*{12}}
\put(1080,143){\circle*{12}}
\put(960,143){\circle*{12}}
\put(841,143){\circle*{12}}
\put(721,154){\circle*{12}}
\put(601,154){\circle*{12}}
\put(481,297){\circle*{12}}
\put(362,324){\circle*{12}}
\put(1339,815){\circle*{12}}
\sbox{\plotpoint}{\rule[-0.200pt]{0.400pt}{0.400pt}}%
\put(1259,765){\makebox(0,0)[r]{Latin Hypercube}}
\put(1439,143){\circle{18}}
\put(1319,161){\circle{18}}
\put(1200,162){\circle{18}}
\put(1080,162){\circle{18}}
\put(960,162){\circle{18}}
\put(841,146){\circle{18}}
\put(721,161){\circle{18}}
\put(601,342){\circle{18}}
\put(481,380){\circle{18}}
\put(362,860){\circle{18}}
\put(1339,765){\circle{18}}
\sbox{\plotpoint}{\rule[-0.500pt]{1.000pt}{1.000pt}}%
\put(1259,715){\makebox(0,0)[r]{Mostlikely Scen.}}
\put(1439,143){\makebox(0,0){$\times$}}
\put(1319,143){\makebox(0,0){$\times$}}
\put(1200,157){\makebox(0,0){$\times$}}
\put(1080,157){\makebox(0,0){$\times$}}
\put(960,157){\makebox(0,0){$\times$}}
\put(841,162){\makebox(0,0){$\times$}}
\put(721,251){\makebox(0,0){$\times$}}
\put(601,297){\makebox(0,0){$\times$}}
\put(481,297){\makebox(0,0){$\times$}}
\put(362,860){\makebox(0,0){$\times$}}
\put(1339,715){\makebox(0,0){$\times$}}
\end{picture}

%% file: t_s.tex
\setlength{\unitlength}{0.240900pt}
\ifx\plotpoint\undefined\newsavebox{\plotpoint}\fi
\sbox{\plotpoint}{\rule[-0.200pt]{0.400pt}{0.400pt}}%
\begin{picture}(1500,900)(0,0)
\font\gnuplot=cmr10 at 10pt
\gnuplot
\sbox{\plotpoint}{\rule[-0.200pt]{0.400pt}{0.400pt}}%
\put(202,143){\makebox(0,0)[r]{0}}
\put(222.0,143.0){\rule[-0.200pt]{4.818pt}{0.400pt}}
\put(202,215){\makebox(0,0)[r]{10}}
\put(222.0,215.0){\rule[-0.200pt]{4.818pt}{0.400pt}}
\put(202,286){\makebox(0,0)[r]{20}}
\put(222.0,286.0){\rule[-0.200pt]{4.818pt}{0.400pt}}
\put(202,358){\makebox(0,0)[r]{30}}
\put(222.0,358.0){\rule[-0.200pt]{4.818pt}{0.400pt}}
\put(202,430){\makebox(0,0)[r]{40}}
\put(222.0,430.0){\rule[-0.200pt]{4.818pt}{0.400pt}}
\put(202,501){\makebox(0,0)[r]{50}}
\put(222.0,501.0){\rule[-0.200pt]{4.818pt}{0.400pt}}
\put(202,573){\makebox(0,0)[r]{60}}
\put(222.0,573.0){\rule[-0.200pt]{4.818pt}{0.400pt}}
\put(202,645){\makebox(0,0)[r]{70}}
\put(222.0,645.0){\rule[-0.200pt]{4.818pt}{0.400pt}}
\put(202,717){\makebox(0,0)[r]{80}}
\put(222.0,717.0){\rule[-0.200pt]{4.818pt}{0.400pt}}
\put(202,788){\makebox(0,0)[r]{90}}
\put(222.0,788.0){\rule[-0.200pt]{4.818pt}{0.400pt}}
\put(202,860){\makebox(0,0)[r]{100}}
\put(222.0,860.0){\rule[-0.200pt]{4.818pt}{0.400pt}}
\put(242,82){\makebox(0,0){0}}
\put(242.0,123.0){\rule[-0.200pt]{0.400pt}{4.818pt}}
\put(362,82){\makebox(0,0){10}}
\put(362.0,123.0){\rule[-0.200pt]{0.400pt}{4.818pt}}
\put(481,82){\makebox(0,0){20}}
\put(481.0,123.0){\rule[-0.200pt]{0.400pt}{4.818pt}}
\put(601,82){\makebox(0,0){30}}
\put(601.0,123.0){\rule[-0.200pt]{0.400pt}{4.818pt}}
\put(721,82){\makebox(0,0){40}}
\put(721.0,123.0){\rule[-0.200pt]{0.400pt}{4.818pt}}
\put(841,82){\makebox(0,0){50}}
\put(841.0,123.0){\rule[-0.200pt]{0.400pt}{4.818pt}}
\put(960,82){\makebox(0,0){60}}
\put(960.0,123.0){\rule[-0.200pt]{0.400pt}{4.818pt}}
\put(1080,82){\makebox(0,0){70}}
\put(1080.0,123.0){\rule[-0.200pt]{0.400pt}{4.818pt}}
\put(1200,82){\makebox(0,0){80}}
\put(1200.0,123.0){\rule[-0.200pt]{0.400pt}{4.818pt}}
\put(1319,82){\makebox(0,0){90}}
\put(1319.0,123.0){\rule[-0.200pt]{0.400pt}{4.818pt}}
\put(1439,82){\makebox(0,0){100}}
\put(1439.0,123.0){\rule[-0.200pt]{0.400pt}{4.818pt}}
\put(242.0,143.0){\rule[-0.200pt]{288.357pt}{0.400pt}}
\put(1439.0,143.0){\rule[-0.200pt]{0.400pt}{172.725pt}}
\put(242.0,860.0){\rule[-0.200pt]{288.357pt}{0.400pt}}
\put(40,501){\makebox(0,0){Error (\%)}}
\put(840,21){\makebox(0,0){Number of Scenarios (\%)}}
\put(242.0,143.0){\rule[-0.200pt]{0.400pt}{172.725pt}}
\sbox{\plotpoint}{\rule[-0.500pt]{1.000pt}{1.000pt}}%
\put(1259,815){\makebox(0,0)[r]{Dupacova Alg.}}
\put(1427,143){\circle*{12}}
\put(1379,143){\circle*{12}}
\put(1319,143){\circle*{12}}
\put(1259,143){\circle*{12}}
\put(1200,143){\circle*{12}}
\put(1140,143){\circle*{12}}
\put(1080,143){\circle*{12}}
\put(1020,143){\circle*{12}}
\put(960,143){\circle*{12}}
\put(900,143){\circle*{12}}
\put(841,143){\circle*{12}}
\put(781,143){\circle*{12}}
\put(721,143){\circle*{12}}
\put(661,143){\circle*{12}}
\put(601,143){\circle*{12}}
\put(541,143){\circle*{12}}
\put(481,143){\circle*{12}}
\put(422,143){\circle*{12}}
\put(362,143){\circle*{12}}
\put(302,143){\circle*{12}}
\put(254,860){\circle*{12}}
\put(1339,815){\circle*{12}}
\sbox{\plotpoint}{\rule[-0.200pt]{0.400pt}{0.400pt}}%
\put(1259,765){\makebox(0,0)[r]{Latin Hypercube}}
\put(1427,143){\circle{18}}
\put(1379,143){\circle{18}}
\put(1319,143){\circle{18}}
\put(1259,143){\circle{18}}
\put(1200,143){\circle{18}}
\put(1140,143){\circle{18}}
\put(1080,143){\circle{18}}
\put(1020,143){\circle{18}}
\put(960,143){\circle{18}}
\put(900,143){\circle{18}}
\put(841,143){\circle{18}}
\put(781,143){\circle{18}}
\put(721,207){\circle{18}}
\put(661,143){\circle{18}}
\put(601,143){\circle{18}}
\put(541,207){\circle{18}}
\put(481,143){\circle{18}}
\put(422,143){\circle{18}}
\put(362,481){\circle{18}}
\put(302,143){\circle{18}}
\put(254,207){\circle{18}}
\put(1339,765){\circle{18}}
\sbox{\plotpoint}{\rule[-0.500pt]{1.000pt}{1.000pt}}%
\put(1259,715){\makebox(0,0)[r]{Mostlikely Scen.}}
\put(1427,143){\makebox(0,0){$\times$}}
\put(1379,143){\makebox(0,0){$\times$}}
\put(1319,143){\makebox(0,0){$\times$}}
\put(1259,143){\makebox(0,0){$\times$}}
\put(1200,143){\makebox(0,0){$\times$}}
\put(1140,143){\makebox(0,0){$\times$}}
\put(1080,143){\makebox(0,0){$\times$}}
\put(1020,143){\makebox(0,0){$\times$}}
\put(960,143){\makebox(0,0){$\times$}}
\put(900,143){\makebox(0,0){$\times$}}
\put(841,143){\makebox(0,0){$\times$}}
\put(781,143){\makebox(0,0){$\times$}}
\put(721,143){\makebox(0,0){$\times$}}
\put(661,143){\makebox(0,0){$\times$}}
\put(601,207){\makebox(0,0){$\times$}}
\put(541,207){\makebox(0,0){$\times$}}
\put(481,207){\makebox(0,0){$\times$}}
\put(422,207){\makebox(0,0){$\times$}}
\put(362,207){\makebox(0,0){$\times$}}
\put(302,207){\makebox(0,0){$\times$}}
\put(254,860){\makebox(0,0){$\times$}}
\put(1339,715){\makebox(0,0){$\times$}}
\end{picture}

%% file: chance.tex
\setlength{\unitlength}{0.240900pt}
\ifx\plotpoint\undefined\newsavebox{\plotpoint}\fi
\sbox{\plotpoint}{\rule[-0.200pt]{0.400pt}{0.400pt}}%
\begin{picture}(1500,900)(0,0)
\sbox{\plotpoint}{\rule[-0.200pt]{0.400pt}{0.400pt}}%
\put(202,860){\makebox(0,0)[r]{100}}
\put(222.0,860.0){\rule[-0.200pt]{4.818pt}{0.400pt}}
\put(202,788){\makebox(0,0)[r]{90}}
\put(222.0,788.0){\rule[-0.200pt]{4.818pt}{0.400pt}}
\put(202,717){\makebox(0,0)[r]{80}}
\put(222.0,717.0){\rule[-0.200pt]{4.818pt}{0.400pt}}
\put(202,645){\makebox(0,0)[r]{70}}
\put(222.0,645.0){\rule[-0.200pt]{4.818pt}{0.400pt}}
\put(202,573){\makebox(0,0)[r]{60}}
\put(222.0,573.0){\rule[-0.200pt]{4.818pt}{0.400pt}}
\put(202,502){\makebox(0,0)[r]{50}}
\put(222.0,502.0){\rule[-0.200pt]{4.818pt}{0.400pt}}
\put(202,430){\makebox(0,0)[r]{40}}
\put(222.0,430.0){\rule[-0.200pt]{4.818pt}{0.400pt}}
\put(202,358){\makebox(0,0)[r]{30}}
\put(222.0,358.0){\rule[-0.200pt]{4.818pt}{0.400pt}}
\put(202,286){\makebox(0,0)[r]{20}}
\put(222.0,286.0){\rule[-0.200pt]{4.818pt}{0.400pt}}
\put(202,215){\makebox(0,0)[r]{10}}
\put(222.0,215.0){\rule[-0.200pt]{4.818pt}{0.400pt}}
\put(202,143){\makebox(0,0)[r]{0}}
\put(222.0,143.0){\rule[-0.200pt]{4.818pt}{0.400pt}}
\put(1439,82){\makebox(0,0){100}}
\put(1439.0,123.0){\rule[-0.200pt]{0.400pt}{4.818pt}}
\put(1319,82){\makebox(0,0){90}}
\put(1319.0,123.0){\rule[-0.200pt]{0.400pt}{4.818pt}}
\put(1200,82){\makebox(0,0){80}}
\put(1200.0,123.0){\rule[-0.200pt]{0.400pt}{4.818pt}}
\put(1080,82){\makebox(0,0){70}}
\put(1080.0,123.0){\rule[-0.200pt]{0.400pt}{4.818pt}}
\put(960,82){\makebox(0,0){60}}
\put(960.0,123.0){\rule[-0.200pt]{0.400pt}{4.818pt}}
\put(841,82){\makebox(0,0){50}}
\put(841.0,123.0){\rule[-0.200pt]{0.400pt}{4.818pt}}
\put(721,82){\makebox(0,0){40}}
\put(721.0,123.0){\rule[-0.200pt]{0.400pt}{4.818pt}}
\put(601,82){\makebox(0,0){30}}
\put(601.0,123.0){\rule[-0.200pt]{0.400pt}{4.818pt}}
\put(481,82){\makebox(0,0){20}}
\put(481.0,123.0){\rule[-0.200pt]{0.400pt}{4.818pt}}
\put(362,82){\makebox(0,0){10}}
\put(362.0,123.0){\rule[-0.200pt]{0.400pt}{4.818pt}}
\put(242,82){\makebox(0,0){0}}
\put(242.0,123.0){\rule[-0.200pt]{0.400pt}{4.818pt}}
\put(242.0,143.0){\rule[-0.200pt]{288.357pt}{0.400pt}}
\put(1439.0,143.0){\rule[-0.200pt]{0.400pt}{172.725pt}}
\put(242.0,860.0){\rule[-0.200pt]{288.357pt}{0.400pt}}
\put(242.0,143.0){\rule[-0.200pt]{0.400pt}{172.725pt}}
\put(40,501){\makebox(0,0){Error (\%)}}
\put(840,21){\makebox(0,0){Number of Scenarios (\%)}}
\sbox{\plotpoint}{\rule[-0.500pt]{1.000pt}{1.000pt}}%
\sbox{\plotpoint}{\rule[-0.200pt]{0.400pt}{0.400pt}}%
\put(1259,815){\makebox(0,0)[r]{Dupacova Alg.}}
\sbox{\plotpoint}{\rule[-0.500pt]{1.000pt}{1.000pt}}%
\put(1439,143){\circle*{12}}
\put(1339,396){\circle*{12}}
\put(1239,396){\circle*{12}}
\put(1140,376){\circle*{12}}
\put(1040,376){\circle*{12}}
\put(940,376){\circle*{12}}
\put(741,860){\circle*{12}}
\put(641,860){\circle*{12}}
\put(541,860){\circle*{12}}
\put(442,860){\circle*{12}}
\put(342,860){\circle*{12}}
\put(1339,815){\circle*{12}}
\sbox{\plotpoint}{\rule[-0.200pt]{0.400pt}{0.400pt}}%
\put(1259,765){\makebox(0,0)[r]{Latin Hypercube}}
\put(1439,143){\circle{18}}
\put(1239,376){\circle{18}}
\put(1140,396){\circle{18}}
\put(1040,376){\circle{18}}
\put(940,396){\circle{18}}
\put(741,396){\circle{18}}
\put(641,396){\circle{18}}
\put(541,376){\circle{18}}
\put(442,266){\circle{18}}
\put(342,545){\circle{18}}
\put(1339,765){\circle{18}}
\sbox{\plotpoint}{\rule[-0.500pt]{1.000pt}{1.000pt}}%
\sbox{\plotpoint}{\rule[-0.200pt]{0.400pt}{0.400pt}}%
\put(1259,715){\makebox(0,0)[r]{Mostlikely Scen.}}
\sbox{\plotpoint}{\rule[-0.500pt]{1.000pt}{1.000pt}}%
\put(1439,143){\makebox(0,0){$\times$}}
\put(1339,376){\makebox(0,0){$\times$}}
\put(1239,376){\makebox(0,0){$\times$}}
\put(1140,376){\makebox(0,0){$\times$}}
\put(1040,376){\makebox(0,0){$\times$}}
\put(940,376){\makebox(0,0){$\times$}}
\put(841,376){\makebox(0,0){$\times$}}
\put(741,376){\makebox(0,0){$\times$}}
\put(641,396){\makebox(0,0){$\times$}}
\put(541,143){\makebox(0,0){$\times$}}
\put(442,860){\makebox(0,0){$\times$}}
\put(342,143){\makebox(0,0){$\times$}}
\put(1339,715){\makebox(0,0){$\times$}}
\sbox{\plotpoint}{\rule[-0.200pt]{0.400pt}{0.400pt}}%
\put(242.0,143.0){\rule[-0.200pt]{288.357pt}{0.400pt}}
\put(1439.0,143.0){\rule[-0.200pt]{0.400pt}{172.725pt}}
\put(242.0,860.0){\rule[-0.200pt]{288.357pt}{0.400pt}}
\put(242.0,143.0){\rule[-0.200pt]{0.400pt}{172.725pt}}
\end{picture}

%% file: risk.tex
\setlength{\unitlength}{0.240900pt}
\ifx\plotpoint\undefined\newsavebox{\plotpoint}\fi
\sbox{\plotpoint}{\rule[-0.200pt]{0.400pt}{0.400pt}}%
\begin{picture}(1500,900)(0,0)
\sbox{\plotpoint}{\rule[-0.200pt]{0.400pt}{0.400pt}}%
\put(182,860){\makebox(0,0)[r]{400}}
\put(202.0,860.0){\rule[-0.200pt]{4.818pt}{0.400pt}}
\put(182,755){\makebox(0,0)[r]{390}}
\put(202.0,755.0){\rule[-0.200pt]{4.818pt}{0.400pt}}
\put(182,650){\makebox(0,0)[r]{380}}
\put(202.0,650.0){\rule[-0.200pt]{4.818pt}{0.400pt}}
\put(182,544){\makebox(0,0)[r]{370}}
\put(202.0,544.0){\rule[-0.200pt]{4.818pt}{0.400pt}}
\put(182,439){\makebox(0,0)[r]{360}}
\put(202.0,439.0){\rule[-0.200pt]{4.818pt}{0.400pt}}
\put(182,334){\makebox(0,0)[r]{350}}
\put(202.0,334.0){\rule[-0.200pt]{4.818pt}{0.400pt}}
\put(1297,273){\makebox(0,0){4.0}}
\put(1297.0,314.0){\rule[-0.200pt]{0.400pt}{4.818pt}}
\put(1163,273){\makebox(0,0){3.5}}
\put(1163.0,314.0){\rule[-0.200pt]{0.400pt}{4.818pt}}
\put(1028,273){\makebox(0,0){3.0}}
\put(1028.0,314.0){\rule[-0.200pt]{0.400pt}{4.818pt}}
\put(894,273){\makebox(0,0){2.5}}
\put(894.0,314.0){\rule[-0.200pt]{0.400pt}{4.818pt}}
\put(760,273){\makebox(0,0){2.0}}
\put(760.0,314.0){\rule[-0.200pt]{0.400pt}{4.818pt}}
\put(625,273){\makebox(0,0){1.5}}
\put(625.0,314.0){\rule[-0.200pt]{0.400pt}{4.818pt}}
\put(491,273){\makebox(0,0){1.0}}
\put(491.0,314.0){\rule[-0.200pt]{0.400pt}{4.818pt}}
\put(356,273){\makebox(0,0){0.5}}
\put(356.0,314.0){\rule[-0.200pt]{0.400pt}{4.818pt}}
\put(222,273){\makebox(0,0){0.0}}
\put(222.0,314.0){\rule[-0.200pt]{0.400pt}{4.818pt}}
\put(1337,860){\makebox(0,0)[l]{25}}
\put(1297.0,860.0){\rule[-0.200pt]{4.818pt}{0.400pt}}
\put(1337,755){\makebox(0,0)[l]{20}}
\put(1297.0,755.0){\rule[-0.200pt]{4.818pt}{0.400pt}}
\put(1337,650){\makebox(0,0)[l]{15}}
\put(1297.0,650.0){\rule[-0.200pt]{4.818pt}{0.400pt}}
\put(1337,544){\makebox(0,0)[l]{10}}
\put(1297.0,544.0){\rule[-0.200pt]{4.818pt}{0.400pt}}
\put(1337,439){\makebox(0,0)[l]{5}}
\put(1297.0,439.0){\rule[-0.200pt]{4.818pt}{0.400pt}}
\put(1337,334){\makebox(0,0)[l]{0}}
\put(1297.0,334.0){\rule[-0.200pt]{4.818pt}{0.400pt}}
\put(222.0,334.0){\rule[-0.200pt]{258.967pt}{0.400pt}}
\put(1297.0,334.0){\rule[-0.200pt]{0.400pt}{126.713pt}}
\put(222.0,860.0){\rule[-0.200pt]{258.967pt}{0.400pt}}
\put(222.0,334.0){\rule[-0.200pt]{0.400pt}{126.713pt}}
\put(40,597){\makebox(0,0){$y_1$}}
\put(1457,597){\makebox(0,0){$y_2$}}
\put(759,212){\makebox(0,0){$\lambda$}}
\put(1051,125){\makebox(0,0)[r]{Obj.Fun Value (on $y_1$-axis)}}
\put(1071.0,125.0){\rule[-0.200pt]{28.908pt}{0.400pt}}
\put(222,351){\usebox{\plotpoint}}
\multiput(222.00,351.58)(0.549,0.499){119}{\rule{0.539pt}{0.120pt}}
\multiput(222.00,350.17)(65.881,61.000){2}{\rule{0.270pt}{0.400pt}}
\multiput(289.00,412.58)(0.577,0.499){113}{\rule{0.562pt}{0.120pt}}
\multiput(289.00,411.17)(65.833,58.000){2}{\rule{0.281pt}{0.400pt}}
\multiput(356.00,470.58)(0.618,0.499){107}{\rule{0.595pt}{0.120pt}}
\multiput(356.00,469.17)(66.766,55.000){2}{\rule{0.297pt}{0.400pt}}
\multiput(424.00,525.58)(0.632,0.498){103}{\rule{0.606pt}{0.120pt}}
\multiput(424.00,524.17)(65.743,53.000){2}{\rule{0.303pt}{0.400pt}}
\multiput(491.00,578.58)(0.657,0.498){99}{\rule{0.625pt}{0.120pt}}
\multiput(491.00,577.17)(65.702,51.000){2}{\rule{0.313pt}{0.400pt}}
\multiput(558.00,629.58)(0.840,0.498){77}{\rule{0.770pt}{0.120pt}}
\multiput(558.00,628.17)(65.402,40.000){2}{\rule{0.385pt}{0.400pt}}
\multiput(625.00,669.58)(1.020,0.497){63}{\rule{0.912pt}{0.120pt}}
\multiput(625.00,668.17)(65.107,33.000){2}{\rule{0.456pt}{0.400pt}}
\multiput(692.00,702.58)(1.222,0.497){53}{\rule{1.071pt}{0.120pt}}
\multiput(692.00,701.17)(65.776,28.000){2}{\rule{0.536pt}{0.400pt}}
\multiput(760.00,730.58)(1.408,0.496){45}{\rule{1.217pt}{0.120pt}}
\multiput(760.00,729.17)(64.475,24.000){2}{\rule{0.608pt}{0.400pt}}
\multiput(827.00,754.58)(1.786,0.495){35}{\rule{1.511pt}{0.119pt}}
\multiput(827.00,753.17)(63.865,19.000){2}{\rule{0.755pt}{0.400pt}}
\multiput(894.00,773.58)(2.001,0.495){31}{\rule{1.676pt}{0.119pt}}
\multiput(894.00,772.17)(63.520,17.000){2}{\rule{0.838pt}{0.400pt}}
\multiput(961.00,790.58)(1.786,0.495){35}{\rule{1.511pt}{0.119pt}}
\multiput(961.00,789.17)(63.865,19.000){2}{\rule{0.755pt}{0.400pt}}
\multiput(1028.00,809.58)(2.866,0.492){21}{\rule{2.333pt}{0.119pt}}
\multiput(1028.00,808.17)(62.157,12.000){2}{\rule{1.167pt}{0.400pt}}
\multiput(1095.00,821.58)(2.480,0.494){25}{\rule{2.043pt}{0.119pt}}
\multiput(1095.00,820.17)(63.760,14.000){2}{\rule{1.021pt}{0.400pt}}
\multiput(1163.00,835.59)(4.390,0.488){13}{\rule{3.450pt}{0.117pt}}
\multiput(1163.00,834.17)(59.839,8.000){2}{\rule{1.725pt}{0.400pt}}
\multiput(1230.00,843.59)(3.873,0.489){15}{\rule{3.078pt}{0.118pt}}
\multiput(1230.00,842.17)(60.612,9.000){2}{\rule{1.539pt}{0.400pt}}
\put(222,351){\circle{18}}
\put(289,412){\circle{18}}
\put(356,470){\circle{18}}
\put(424,525){\circle{18}}
\put(491,578){\circle{18}}
\put(558,629){\circle{18}}
\put(625,669){\circle{18}}
\put(692,702){\circle{18}}
\put(760,730){\circle{18}}
\put(827,754){\circle{18}}
\put(894,773){\circle{18}}
\put(961,790){\circle{18}}
\put(1028,809){\circle{18}}
\put(1095,821){\circle{18}}
\put(1163,835){\circle{18}}
\put(1230,843){\circle{18}}
\put(1297,852){\circle{18}}
\put(1131,125){\circle{18}}
\put(1051,75){\makebox(0,0)[r]{Prod/Inv Cost (on $y_1$-axis)}}
\put(1071.0,75.0){\rule[-0.200pt]{28.908pt}{0.400pt}}
\put(222,351){\usebox{\plotpoint}}
\multiput(289.00,351.59)(7.389,0.477){7}{\rule{5.460pt}{0.115pt}}
\multiput(289.00,350.17)(55.667,5.000){2}{\rule{2.730pt}{0.400pt}}
\multiput(356.00,356.58)(3.518,0.491){17}{\rule{2.820pt}{0.118pt}}
\multiput(356.00,355.17)(62.147,10.000){2}{\rule{1.410pt}{0.400pt}}
\put(222.0,351.0){\rule[-0.200pt]{16.140pt}{0.400pt}}
\multiput(491.00,366.58)(1.086,0.497){59}{\rule{0.965pt}{0.120pt}}
\multiput(491.00,365.17)(64.998,31.000){2}{\rule{0.482pt}{0.400pt}}
\multiput(558.00,397.58)(0.632,0.498){103}{\rule{0.606pt}{0.120pt}}
\multiput(558.00,396.17)(65.743,53.000){2}{\rule{0.303pt}{0.400pt}}
\multiput(625.00,450.58)(0.799,0.498){81}{\rule{0.738pt}{0.120pt}}
\multiput(625.00,449.17)(65.468,42.000){2}{\rule{0.369pt}{0.400pt}}
\multiput(692.00,492.58)(0.811,0.498){81}{\rule{0.748pt}{0.120pt}}
\multiput(692.00,491.17)(66.448,42.000){2}{\rule{0.374pt}{0.400pt}}
\multiput(760.00,534.58)(1.613,0.496){39}{\rule{1.376pt}{0.119pt}}
\multiput(760.00,533.17)(64.144,21.000){2}{\rule{0.688pt}{0.400pt}}
\multiput(827.00,555.58)(1.086,0.497){59}{\rule{0.965pt}{0.120pt}}
\multiput(827.00,554.17)(64.998,31.000){2}{\rule{0.482pt}{0.400pt}}
\multiput(894.00,586.58)(1.538,0.496){41}{\rule{1.318pt}{0.120pt}}
\multiput(894.00,585.17)(64.264,22.000){2}{\rule{0.659pt}{0.400pt}}
\multiput(961.00,608.58)(1.086,0.497){59}{\rule{0.965pt}{0.120pt}}
\multiput(961.00,607.17)(64.998,31.000){2}{\rule{0.482pt}{0.400pt}}
\multiput(1028.00,639.58)(2.130,0.494){29}{\rule{1.775pt}{0.119pt}}
\multiput(1028.00,638.17)(63.316,16.000){2}{\rule{0.888pt}{0.400pt}}
\multiput(1095.00,655.58)(1.222,0.497){53}{\rule{1.071pt}{0.120pt}}
\multiput(1095.00,654.17)(65.776,28.000){2}{\rule{0.536pt}{0.400pt}}
\multiput(1163.00,683.58)(2.866,0.492){21}{\rule{2.333pt}{0.119pt}}
\multiput(1163.00,682.17)(62.157,12.000){2}{\rule{1.167pt}{0.400pt}}
\multiput(1230.00,695.58)(2.443,0.494){25}{\rule{2.014pt}{0.119pt}}
\multiput(1230.00,694.17)(62.819,14.000){2}{\rule{1.007pt}{0.400pt}}
\put(222,351){\circle*{12}}
\put(289,351){\circle*{12}}
\put(356,356){\circle*{12}}
\put(424,366){\circle*{12}}
\put(491,366){\circle*{12}}
\put(558,397){\circle*{12}}
\put(625,450){\circle*{12}}
\put(692,492){\circle*{12}}
\put(760,534){\circle*{12}}
\put(827,555){\circle*{12}}
\put(894,586){\circle*{12}}
\put(961,608){\circle*{12}}
\put(1028,639){\circle*{12}}
\put(1095,655){\circle*{12}}
\put(1163,683){\circle*{12}}
\put(1230,695){\circle*{12}}
\put(1297,709){\circle*{12}}
\put(1131,75){\circle*{12}}
\put(424.0,366.0){\rule[-0.200pt]{16.140pt}{0.400pt}}
\put(1051,25){\makebox(0,0)[r]{Risk (on $y_2$-axis)}}
\put(1071.0,25.0){\rule[-0.200pt]{28.908pt}{0.400pt}}
\put(222,833){\usebox{\plotpoint}}
\multiput(222.00,831.92)(2.638,-0.493){23}{\rule{2.162pt}{0.119pt}}
\multiput(222.00,832.17)(62.514,-13.000){2}{\rule{1.081pt}{0.400pt}}
\multiput(289.00,818.92)(1.204,-0.497){53}{\rule{1.057pt}{0.120pt}}
\multiput(289.00,819.17)(64.806,-28.000){2}{\rule{0.529pt}{0.400pt}}
\multiput(356.00,790.92)(1.035,-0.497){63}{\rule{0.924pt}{0.120pt}}
\multiput(356.00,791.17)(66.082,-33.000){2}{\rule{0.462pt}{0.400pt}}
\put(424,757.67){\rule{16.140pt}{0.400pt}}
\multiput(424.00,758.17)(33.500,-1.000){2}{\rule{8.070pt}{0.400pt}}
\multiput(491.00,756.92)(0.620,-0.498){105}{\rule{0.596pt}{0.120pt}}
\multiput(491.00,757.17)(65.762,-54.000){2}{\rule{0.298pt}{0.400pt}}
\multiput(558.58,701.65)(0.499,-0.582){131}{\rule{0.120pt}{0.566pt}}
\multiput(557.17,702.83)(67.000,-76.826){2}{\rule{0.400pt}{0.283pt}}
\multiput(625.00,624.92)(0.644,-0.498){101}{\rule{0.615pt}{0.120pt}}
\multiput(625.00,625.17)(65.723,-52.000){2}{\rule{0.308pt}{0.400pt}}
\multiput(692.00,572.92)(0.774,-0.498){85}{\rule{0.718pt}{0.120pt}}
\multiput(692.00,573.17)(66.509,-44.000){2}{\rule{0.359pt}{0.400pt}}
\multiput(760.00,528.92)(1.786,-0.495){35}{\rule{1.511pt}{0.119pt}}
\multiput(760.00,529.17)(63.865,-19.000){2}{\rule{0.755pt}{0.400pt}}
\multiput(827.00,509.92)(1.204,-0.497){53}{\rule{1.057pt}{0.120pt}}
\multiput(827.00,510.17)(64.806,-28.000){2}{\rule{0.529pt}{0.400pt}}
\multiput(894.00,481.92)(2.130,-0.494){29}{\rule{1.775pt}{0.119pt}}
\multiput(894.00,482.17)(63.316,-16.000){2}{\rule{0.888pt}{0.400pt}}
\multiput(961.00,465.92)(1.695,-0.496){37}{\rule{1.440pt}{0.119pt}}
\multiput(961.00,466.17)(64.011,-20.000){2}{\rule{0.720pt}{0.400pt}}
\multiput(1028.00,445.92)(3.137,-0.492){19}{\rule{2.536pt}{0.118pt}}
\multiput(1028.00,446.17)(61.736,-11.000){2}{\rule{1.268pt}{0.400pt}}
\multiput(1095.00,434.92)(2.310,-0.494){27}{\rule{1.913pt}{0.119pt}}
\multiput(1095.00,435.17)(64.029,-15.000){2}{\rule{0.957pt}{0.400pt}}
\multiput(1163.00,419.93)(4.390,-0.488){13}{\rule{3.450pt}{0.117pt}}
\multiput(1163.00,420.17)(59.839,-8.000){2}{\rule{1.725pt}{0.400pt}}
\multiput(1230.00,411.93)(4.390,-0.488){13}{\rule{3.450pt}{0.117pt}}
\multiput(1230.00,412.17)(59.839,-8.000){2}{\rule{1.725pt}{0.400pt}}
\put(222,833){\makebox(0,0){$\times$}}
\put(289,820){\makebox(0,0){$\times$}}
\put(356,792){\makebox(0,0){$\times$}}
\put(424,759){\makebox(0,0){$\times$}}
\put(491,758){\makebox(0,0){$\times$}}
\put(558,704){\makebox(0,0){$\times$}}
\put(625,626){\makebox(0,0){$\times$}}
\put(692,574){\makebox(0,0){$\times$}}
\put(760,530){\makebox(0,0){$\times$}}
\put(827,511){\makebox(0,0){$\times$}}
\put(894,483){\makebox(0,0){$\times$}}
\put(961,467){\makebox(0,0){$\times$}}
\put(1028,447){\makebox(0,0){$\times$}}
\put(1095,436){\makebox(0,0){$\times$}}
\put(1163,421){\makebox(0,0){$\times$}}
\put(1230,413){\makebox(0,0){$\times$}}
\put(1297,405){\makebox(0,0){$\times$}}
\put(1131,25){\makebox(0,0){$\times$}}
\put(222.0,334.0){\rule[-0.200pt]{258.967pt}{0.400pt}}
\put(1297.0,334.0){\rule[-0.200pt]{0.400pt}{126.713pt}}
\put(222.0,860.0){\rule[-0.200pt]{258.967pt}{0.400pt}}
\put(222.0,334.0){\rule[-0.200pt]{0.400pt}{126.713pt}}
\end{picture}